\documentclass[]{jingdong}
\usepackage[toc,page,header]{appendix}

\usepackage{latexsym}
\usepackage[T1]{fontenc}
\usepackage[utf8]{inputenc}
\usepackage{microtype}
\usepackage{inconsolata}
\usepackage{graphicx}
\usepackage{hyperref}       
\usepackage{url}            
\usepackage{booktabs}       
\usepackage{amsfonts}       
\usepackage{nicefrac}       
\usepackage{stackengine}
\usepackage{microtype}      
\usepackage{colortbl}
\usepackage{xcolor}
\usepackage{amsmath}
\usepackage{amssymb}
\usepackage{amsthm}
\usepackage{mathrsfs}
\usepackage{pifont}
\usepackage{MnSymbol}
\usepackage{balance}
\usepackage{enumitem}
\usepackage{listings}
\usepackage{xcolor}
\usepackage{natbib}
\usepackage{multicol}

\AtBeginDocument{%
  \providecommand\BibTeX{{%
    \normalfont B\kern-0.5em{\scshape i\kern-0.25em b}\kern-0.8em\TeX}}}

\makeatletter
\DeclareRobustCommand\onedot{\futurelet\@let@token\@onedot}
\def\@onedot{\ifx\@let@token.\else.\null\fi}
\newcommand{\eg}{\emph{e.g\@\onedot}}

\usepackage{setspace}
\usepackage{mathtools}

\usepackage{multirow,booktabs}
\usepackage{subcaption}

\newcommand{\owo}[1]{\textsc{OAgents}}

\definecolor{lightgreen}{RGB}{144, 238, 144} 
\definecolor{lightred}{RGB}{255, 105, 97}

\newtcolorbox{promptbox}[2][Prompt]{
colback=black!5!white,
arc=5pt, 
boxrule=0.5pt,
fonttitle=\bfseries,
title=#1, 
before upper={\small}, fontupper=\fontfamily{ptm}\selectfont,
colframe=#2, 
}
\definecolor{ogreen}{RGB}{34, 139, 34}
\microtypesetup{expansion=false}
\usepackage{tabularx}
\usepackage{subcaption}
\usepackage[textsize=tiny]{todonotes}
\usepackage{fontawesome}
\usepackage[most]{tcolorbox}
\usepackage{enumitem}
\usepackage{makecell}

\usepackage{xspace}
\newcommand{\method}{Echo-Memory\xspace}

\definecolor{themepink}{RGB}{190,82,116}
\newcommand{\themebf}[1]{{\color{themepink}\textbf{#1}}}
\newcommand{\tableheader}{\rowcolor{red!6}}

\providecommand{\eg}{{e.g.}}

\newcommand{\thinparagraph}[1]{\vspace{0.1em}\noindent\textbf{#1}\enspace}

\newcommand{\finding}[2]{%
    \vspace{0.25em}
    \begin{tcolorbox}[
        colback=blue!3!white,
        colframe=blue!55!black,
        arc=5pt,
        boxsep=5pt,
        left=10pt, right=10pt, top=4pt, bottom=4pt,
        boxrule=0.8pt,
        drop shadow=gray!35!white,
        enhanced jigsaw
    ]
        \paragraph{\textbf{\textit{Finding #1:}}} #2
    \end{tcolorbox}
    \vspace{0.35em}
}

\title{Echo-Memory: A Controlled Study of Memory in Action World Models}

\renewcommand{\authorlist}{%
{\normalfont\bfseries\fontsize{11}{13}\selectfont
Wayne~King$^{1}$,
Zeyue~Xue$^{2,*}$,
Yuxuan~Bian$^{3}$,
Jie~Huang$^{2}$,
Haoran~Li$^{2}$,
Yaowei~Li$^{4}$,
Yaofeng~Su$^{5}$,
Yuming~Li$^{2}$,
Haoyu~Wang$^{6}$,
Shiyi~Zhang$^{2}$,
Songchun~Zhang$^{7}$,
Yuwei~Niu$^{4}$,
Sihan~Xu$^{8}$,
Junhao~Zhuang$^{6}$,
Haoyang~Huang$^{2}$,
Nan~Duan$^{2,\dagger}$}}
\renewcommand{\affiliationlist}{%
{\normalfont\fontsize{9}{10.5}\selectfont
$^{1}$The University of Hong Kong;
$^{2}$Joy Future Academy, JD;
$^{3}$The Chinese University of Hong Kong;
$^{4}$Peking University;
$^{5}$Fudan University;
$^{6}$Tsinghua University;
$^{7}$The Hong Kong University of Science and Technology;
$^{8}$University of Michigan.}}
\renewcommand{\contributionlist}{%
{\normalfont\fontsize{9}{10.5}\selectfont $^{*}$Project leader. $^{\dagger}$Project advisor.}}

\abstract{%
We present \textbf{Echo-Memory}, a controlled study of memory
mechanisms in action-conditioned world models. These models generate
multi-segment videos from a first frame, text prompt, and camera-action
sequence, but their central failure is often memory rather than local
image synthesis: after the camera leaves and returns, the scene or
salient object may silently change. Existing memory designs are hard to
compare because gains are entangled with backbone, training, retrieval,
and evaluation differences.
Echo-Memory fixes the action-to-video interface and varies only how
history is stored and read by the generator. Under a shared video
diffusion backbone, optimizer, camera-action representation, sampler,
and evaluation pipeline, we compare raw context, compression-based
memory, spatial summaries with different read-out paths, and
state-space recurrence. This matched matrix separates four otherwise
conflated axes: \emph{capacity}, \emph{compression}, \emph{read-out},
and \emph{recurrence}.
We also evaluate memory through a three-branch protocol: replay quality,
in-domain loop revisit, and open-domain return probes. The branches
routinely disagree, showing that replay fidelity is not a sufficient
proxy for remembering a world. Three findings follow. Raw context is a
strong capacity baseline and improves open-domain return far more than
it improves replay metrics. Compactness is not a free substitute for
capacity: aggressive spatial and hybrid-compression memories lose the
salient evidence needed for return. Finally, block-wise state-space
recurrence is the strongest open-domain return mechanism in our matrix,
showing that the structure of implicit memory matters as much as the
decision to use it. These results provide a compact protocol for
studying memory in action world models beyond isolated replay metrics.
}

\date{\today}
\checkdata[Project Page]{\url{https://echo-team-joy-future-academy-jd.github.io/Echo-Memory/}}
\checkdata[Code]{\url{https://github.com/Echo-Team-Joy-Future-Academy-JD/Echo-Memory}}

\begin{document}
\maketitle

\section{Introduction}
\label{sec:intro}

A central frontier for video generation is the move from producing a
single plausible clip to producing a controlled \emph{world rollout}.
An action world model receives a first frame, a text prompt, and a
sequence of camera actions. It must generate the next chunk, then the
next, and continue doing so while preserving geometry, object
identity, and camera obedience across revisits. The visible failures
are familiar: the camera returns to the starting pose but the scene
has silently changed, the salient object is replaced by a plausible
impostor, or a chunk boundary wipes away the accumulated context. At
their root, these are not generic generation failures. They are
\emph{memory} failures, and they are the failure mode that decides
whether a system has actually modelled a world rather than merely
extrapolated a clip.

Recent work has proposed several ways to supply this missing memory.
Context-based methods retrieve previous observations and append them
as visual evidence for the current segment~\cite{yu2025context,hong2025relic,wu2025video,lee20253d}.
Compression-based methods reweight or shorten the retrieved history
to reduce cost~\cite{cai2025mixture,wu2025pack,ji2025memflow,zhu2025memorize}.
Spatial memories replace the full temporal stack with compact scene
tokens~\cite{wu2025video,zhao2025spatia,wang2026anchorweave,garcin2026beyond,hu2026geometry},
whereas state-space variants carry history implicitly through
recurrence~\cite{po2025long,yu2025videossm,chen2025recurrent,zhang2025test,lillemark2026flow}.
The same pressure appears in plug-and-play memory tokens, multi-shot
keyframe memories, real-time interactive world models, persistent-world
editors, and efficient long-form video diffusion
systems~\cite{song2025learning,chen2025first,meng2025holocine,an2025onestory,zhang2025storymem,zhou2026videomemory,yang2025longlive,sun2025worldplay,wu2026infinite,po2026multigen,savva2026solaris,chen2026teleworld,mao2025yume,team2026lingbotworld,fsvideo2026fsvideo}.
These studies make clear that memory is now a central design axis for
action-conditioned video generation. They also make comparison
difficult: reported gains are often entangled with changes in backbone,
training recipe, context budget, sampling procedure, and evaluation
protocol.

This paper studies memory from a controlled, vision-centric perspective.
Rather than introducing another memory module, \method{} uses a fixed
video diffusion-transformer backbone, data interface, camera-action
encoding, sampler, and evaluation pipeline, and varies only the memory
representation. This setting turns memory design into a common
interface question: what information is stored, how is it compressed,
how is it read by the generator, and whether it can survive a return
motion after the camera leaves the visible support. The resulting
comparison is intentionally narrower than the space of all world
models, but it exposes differences that are otherwise easy to hide
behind unrelated system changes.

\begin{figure}[t]
\centering
\IfFileExists{fig/figure_1_abs_framework.pdf}{%
    \includegraphics[width=\linewidth]{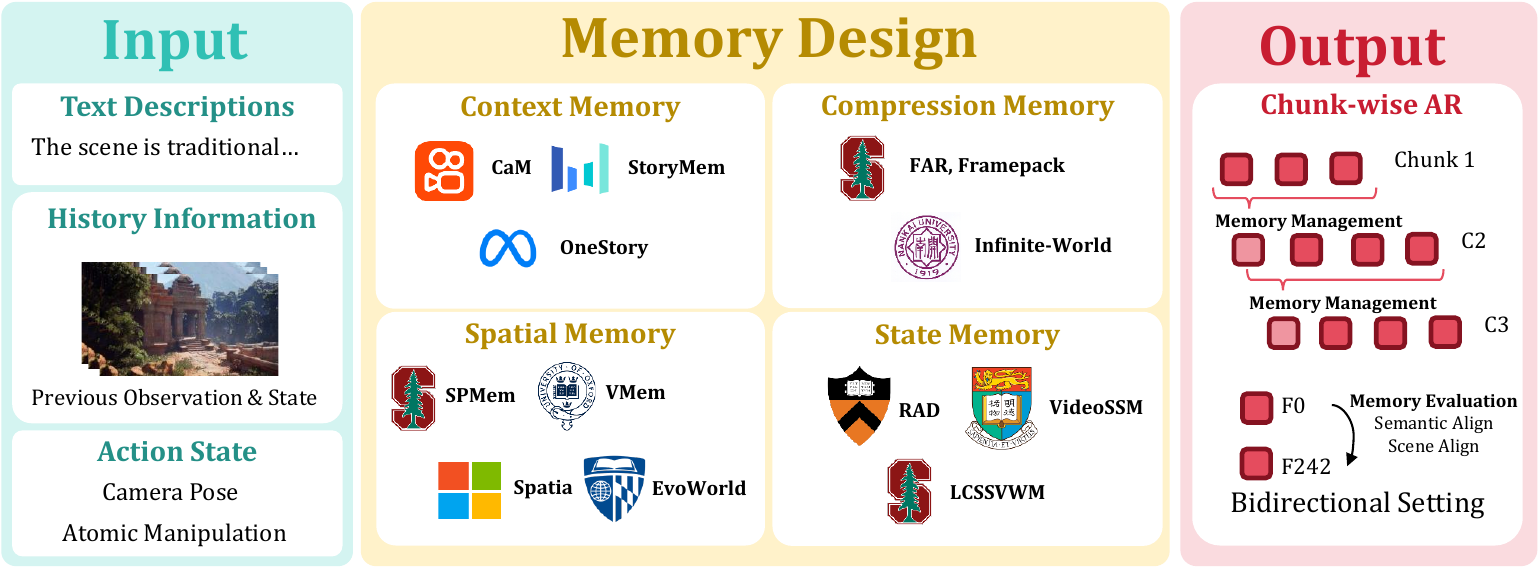}%
}{%
    \fbox{\parbox{0.92\linewidth}{\centering
    \textbf{[Figure placeholder: abstract teaser and workflow]}\\[2pt]
    Input text, history, and action state condition memory design, which
    supports chunk-wise world generation and revisit consistency.}}%
}
\caption{\textbf{Abstract teaser and workflow of \method{}.}
Given a text description, historical observations, and the
camera/action state, an action world model must generate chunk-wise
video while carrying memory across revisits. The figure positions
Context, Compression, Spatial, and State-Space families as
representative designs for preserving a revisitable world.}
\label{fig:conceptual-memory}
\end{figure}

Several methodological concerns motivate this design. Larger raw
context may improve performance simply because it exposes more
evidence, not because it provides a better memory mechanism;
compression can appear efficient when token cost is hidden inside an
operator whose retained evidence is never measured; and in-domain
trajectories can overstate memory quality, because familiar textures
allow plausible local reconstruction even when object identity is lost
under distribution shift. We therefore evaluate every memory row
through the same three branches: ground-truth replay, in-domain loop
closure, and open-domain return. This also forces a separation between
storage and read-out: a memory can store evidence without making it
accessible to the generator, and one of our spatial read-out ablations
shows that this distinction changes replay quality without improving
revisit consistency.

Our study is organized around four memory families and three evaluation
branches. The main empirical results are as follows:
\begin{itemize}[topsep=4pt,itemsep=2pt]
    \item \textbf{Raw context is a strong capacity baseline.} Increasing
    context from $K{=}1$ to $K{=}20$ improves the replay image bundle and
    raises open-domain VLM return from $12.25$ to $58.63$.
    \item \textbf{Compact memory is not automatically semantic memory.}
    Spatial Memory is competitive on replay PSNR but weak on
    open-domain return, while hybrid compression loses much of the
    signal preserved by simpler length compression.
    \item \textbf{Read-out matters as much as storage.} Storing tokens is
    insufficient unless the generator can use them at the moment of
    return; withheld and dedicated spatial read-out variants expose this
    difference.
    \item \textbf{Structured recurrence is the strongest current
    open-domain bias.} Block-wise State-Space memory reaches an
    open-domain VLM score of $69.00$ despite lower replay PSNR,
    illustrating why replay alone is not an adequate memory score.
\end{itemize}

The rest of the main paper gives a compact literature review, defines
the shared interface, describes the three-branch evaluation, and reports
the controlled comparisons. Extended related work, training details, and
the full ablation analysis are provided in the appendix.
\section{Related Work}
\label{sec:related-main}

\thinparagraph{Action world models.}
World models have been studied as latent imagination, predictive state
dynamics, and joint-embedding prediction. Recent video and embodied
systems move this idea into generated visual rollouts conditioned on
prompts, camera motion, or agent actions. In this setting, memory is not
only a cache-management problem: the model must preserve scene geometry,
object identity, and action consistency after the camera leaves and
later revisits a region.

\thinparagraph{Memory mechanisms.}
Existing long-horizon video systems typically use one of three memory
interfaces. Retrieval methods append previous frames or features chosen
by camera overlap, scene geometry, entity identity, or temporal
proximity. Compression methods reduce the cost of history by weighting
tokens, pooling temporal windows, packing frames, or replacing full
frame stacks with compact summaries. Implicit methods carry history
through recurrent or state-space computation instead of explicit visual
tokens. These designs are usually compared under different backbones,
budgets, schedules, and metrics, making it hard to isolate the memory
mechanism itself.

\thinparagraph{Evaluation.}
Standard video metrics such as PSNR, SSIM, LPIPS, FID, and FVD remain
useful health checks, but they do not directly measure revisit
consistency. Recent action-world-model work has begun to report
loop-closure and return metrics, while VLM-as-judge protocols provide a
way to verify semantic scene and identity preservation in open-domain
settings. \method{} combines these ideas in a controlled matrix:
Replay measures camera-following quality, while in-domain and
open-domain return probes measure whether the same world survives after
a leave-and-return motion. A fuller literature review is provided in
\cref{sec:related}.

\section{Method: The \method{} Design Space}
\label{sec:method}

This section formalizes the memory design space studied by
\method{}. The point is not a new backbone, but a clean
factorization of how an action world model can represent the past:
Context tokens, Compression operators, Spatial summaries, or a
State-Space state. All rows share the same training path and the same
evaluation path; the row name changes only the memory/context profile.

\subsection{Preliminaries}
\label{sec:method-prelim}

\thinparagraph{Backbone.}
Let $\mathbf{x}_{1:T}\in\mathbb{R}^{T\times H\times W\times C}$ be a
target video segment of $T$ frames at spatial resolution
$H{\times}W$.  A pre-trained video
diffusion-transformer~\cite{wan2025wan,alibaba2025wan25} models the
conditional velocity field $v_\theta(\mathbf{z}_t; t,
\mathbf{c}_\text{text}, \mathbf{c}_\text{ctx},
\mathbf{c}_\text{act})$, where $\mathbf{z}_t$ is the noised
VAE-encoded latent at flow-matching timestep $t$,
$\mathbf{c}_\text{text}$ is a text embedding,
$\mathbf{c}_\text{ctx}$ is a memory context, and
$\mathbf{c}_\text{act}\in\mathbb{R}^{T\times 12}$ is a per-frame
relative-RT camera-action sequence. The single-stage training
objective is the rectified-flow regression loss applied
\emph{only on the target frames}:
\begin{equation}
  \mathcal{L}(\theta) = \mathbb{E}_{t,\mathbf{x},\mathbf{c}}
    \big\| v_\theta(\mathbf{z}_t; t, \mathbf{c}_\text{text},
       \mathbf{c}_\text{ctx}, \mathbf{c}_\text{act})
       - v^\star \big\|_2^2,
   \qquad v^\star = \mathbf{z}_1 - \mathbf{z}_0 .
   \label{eq:loss}
\end{equation}

\thinparagraph{Per-frame VAE context.}
We use a per-frame VAE context representation, which produces one
latent token group per video frame. This keeps the temporal dimension
of $\mathbf{c}_\text{ctx}$ semantically aligned with the target tokens,
and makes compression-based memory operations well defined as
transformations on the temporal axis of a fixed-length latent stack.

\thinparagraph{Action representation.}
We use a $12$-dimensional relative camera RT with $9$ rotation entries
and $3$ translation entries, expressed in the reference frame of the
first frame of the segment. Every action in this study is
relative; absolute action encodings are out of scope.

\thinparagraph{Context-conditioned generation.}
For a current segment with start frame $s$ and end frame $e$, the
model receives the target segment together with a context set of $K$
historical observations:
\begin{equation}
   \mathcal{C}_{s:e}
   =
   \big\{(\mathbf{x}_{f_k}, \mathbf{p}_{f_k})\big\}_{k=0}^{K-1},
   \quad f_0 = s,\;\;f_{1..K-1}\subset\{0,\ldots,s-1\},
\end{equation}
where $\mathbf{p}_{f_k}$ denotes the associated camera pose/action
metadata. The first context element is the anchor observation for the
current segment; the remaining $K{-}1$ elements come from a retriever
$\mathcal{R}$ over historical frames. With probability
$p_\text{drop}=0.1$ we drop the retrieved frames and keep only the
anchor, simulating a cold-start regime.

\subsection{Memory Design Space}
\label{sec:method-design-space}

\begin{figure}[htbp]
\centering
\IfFileExists{fig/figure_2_mem_overview.pdf}{%
    \includegraphics[width=\linewidth]{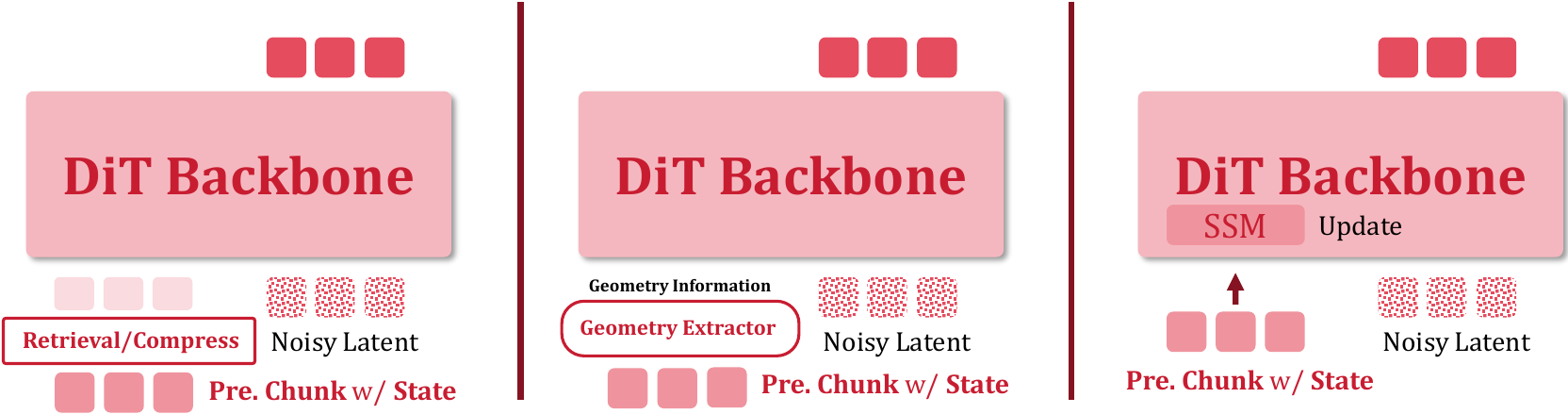}%
}{%
    \fbox{\parbox{0.96\linewidth}{\centering
    \textbf{[Figure placeholder: overview of four memory approaches]}\\[2pt]
    Context and Compression memory retrieve or compact previous chunks;
    Spatial Memory reads a geometric scene summary; State-Space Memory
    propagates history through recurrent state.}}%
}
\caption{\textbf{Overview of four representative approaches to memory
in action world models.} Under a shared video diffusion backbone and a
shared camera-action interface, the approaches differ only in how
historical information is stored and read. We compare four families
throughout: Context, Compression, Spatial, and State-Space.}
\label{fig:memory-mechanisms}
\end{figure}

\method{} treats memory as a pair of operations attached to an
otherwise fixed video DiT and our design can be summerized as in \ref{fig:memory-mechanisms}.

\subsection{Concrete Variants}
\label{sec:method-variants}

The representative rows instantiate the four families above under the
same backbone and action interface: anchor-only I2V, raw Context at
different lengths, Compression, Spatial Memory, and State-Space
recurrence. Focused ablations then vary one mechanism at a time:
Spatial read-out, Compression type, recurrence structure, or raw
context length. The empirical tables introduce the exact rows where
they are needed, so the method section can stay focused on the shared
interface rather than on a separate registry.

\thinparagraph{Default configuration.}
Unless otherwise stated, every row uses relative-RT camera conditioning,
a fixed flow-matching timestep shift, target-frame-only supervision,
and the same inference-time memory profile as its training
configuration.

\subsection{A Unified Memory Interface}
\label{sec:method-registry}

To keep the comparison fair, the training loop, sampler, and evaluation
pipeline expose the same memory interface to every variant. Each row is
specified by a small memory profile: storage type, context length,
compression rule, read-out path, and recurrence structure when
applicable. Mutual-exclusion constraints are enforced at the profile
level; for example, the two State-Space instantiations cannot be
enabled simultaneously, and length compression is kept separate from
pure token weighting unless the row is explicitly marked as hybrid.
This unified interface is what makes Sec.~\ref{sec:ablation} a
controlled comparison rather than a juxtaposition of independently
tuned systems.
\section{Training Protocol}
\label{sec:training-main}

All variants are trained under a single shared protocol. The backbone,
optimizer, schedule, sampler, data interface, action representation, and
evaluation path are fixed; only the memory profile changes. This is the
main control that lets the later tables be read as storage,
compression, read-out, or recurrence effects rather than as hidden
changes in training recipe.

\thinparagraph{Data and actions.}
We train on the Context-as-Memory dataset, which provides long camera
trajectories, per-frame poses, prompts, and source videos. Each target
sample is an $81$-frame segment at $352{\times}640$ resolution. The
camera signal is a per-frame $12$-dimensional relative-RT vector,
expressed in the reference frame of the first target frame. Context
frames are selected by a fixed field-of-view retrieval policy, and the
same reference-frame convention is used during training and evaluation.

\begin{table}[t]
\centering
\small
\caption{Single-stage hyper-parameters shared by all variants. Only
context length and memory module vary across rows.}
\label{tab:main-hparams}
\begin{tabular}{ll}
\toprule
\tableheader
\textbf{Setting} & \textbf{Value} \\
\midrule
Backbone                         & Video DiT (per-frame VAE) \\
Resolution                       & $352\times 640$ \\
Segment length                   & $81$ frames \\
Context length $K$               & $\{1,5,20\}$, default $5$ \\
Memory module                    & Context, Compression, Spatial, or State-Space \\
Optimizer                        & AdamW \\
Learning rate                    & $5\!\times\!10^{-5}$ ($1\!\times\!10^{-4}$ at $K{=}1$) \\
GPUs                             & $8$ A100-80G \\
Total steps                      & $5$k \\
Target-frame-only supervision    & enabled \\
Relative-RT action encoding      & enabled \\
$10\%$ overlap-drop policy       & enabled \\
\bottomrule
\end{tabular}
\end{table}

\thinparagraph{Replay diagnostics.}
During training we sample fixed replay cases at matched intervals.
These samples use the same generation path as evaluation, including the
same context retriever, RT-relative action constructor, and memory
profile. The scalar denoising loss is therefore supplemented by a visual
diagnostic that exposes boundary continuity, identity drift, and
action-visual alignment before final evaluation. Full training and
sampling details are deferred to \cref{sec:training}.

\begin{figure}[t]
\centering
\IfFileExists{fig/replay_progression_panel.pdf}{%
    \includegraphics[width=\linewidth]{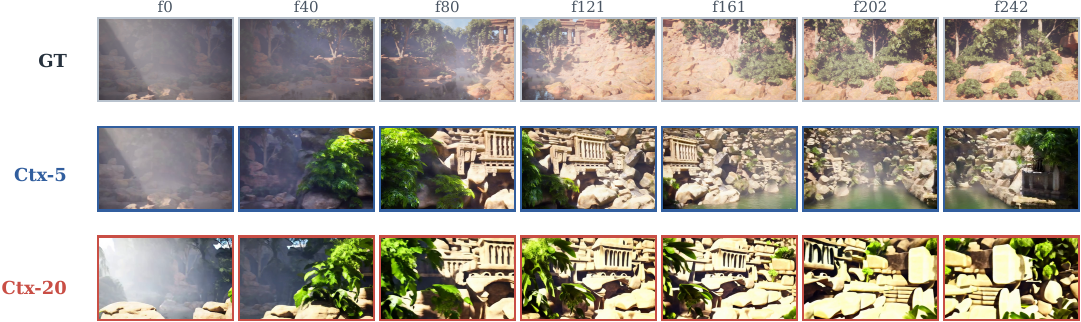}%
}{%
    \fbox{\parbox{0.92\linewidth}{\centering
    \textbf{[Figure placeholder: replay progression panel]}\\[2pt]
    Rows compare GT replay and generated multi-segment replay at matched
    time indices.}}%
}
\caption{\textbf{Replay progression on a fixed GT camera trajectory.}
The diagnostic samples compare generated multi-segment replay against a
dataset trajectory at matched time indices. The panel reveals where
background structure, object layout, and boundary continuity begin to
drift before the final revisit probes are run.}
\label{fig:main-sampling-progress}
\end{figure}

The replay diagnostic is useful precisely because the scalar
flow-matching loss is a weak indicator of memory behavior. Two runs can
have nearly identical denoising loss while diverging sharply at chunk
boundaries or during return motion. Conversely, a model with slightly
worse loss can preserve identity better if its memory path carries
useful evidence across segments. We therefore treat replay sampling as a
fixed in-training protocol rather than as an informal visualization:
the held-out first frames, action constructor, context retriever, and
generation code path are all shared across variants.

This protocol also provides a bridge between training and evaluation.
Replay samples do not replace the final return metrics, because they
still follow dataset-backed camera trajectories. However, they expose
the first point at which a memory mechanism becomes unreliable: a
context row may drift only after several generated chunks, a spatial
row may preserve background layout but lose object identity, and a
state-space row may carry scene evidence while paying a local
reconstruction cost. These qualitative differences motivate the
three-branch evaluation used in the next section.

\section{Evaluation}
\label{sec:evaluation}

\begin{figure}[t]
\centering
\IfFileExists{fig/figure_3_mem_eval.pdf}{%
    \includegraphics[width=0.95\linewidth]{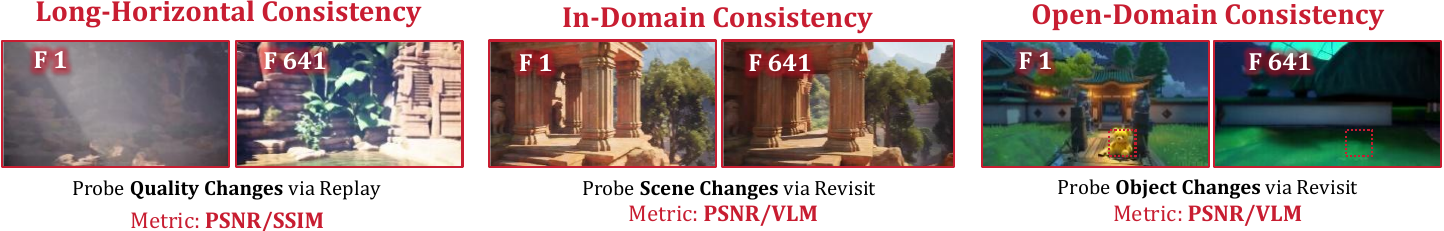}%
}{%
    \fbox{\parbox{0.92\linewidth}{\centering
    \textbf{[Figure placeholder: evaluation taxonomy]}\\[2pt]
    Replay, in-domain loop closure, and open-domain return probing.}}%
}
\caption{Evaluation taxonomy used in the study. Replay measures
long-horizon image quality under GT camera motion. In-domain and
open-domain return probes measure whether visual evidence survives a
leave-and-return trajectory.}
\label{fig:eval-sample-grid}
\end{figure}

The central evaluation challenge is that action world models mix two
capabilities that should not be collapsed into one score. A model must
follow camera actions, but it must also preserve the scene after the
camera leaves and later returns. We therefore report three branches:
\themebf{Replay}, which follows dataset camera trajectories and reports
PSNR/SSIM/LPIPS; \themebf{in-domain return}, which uses a
GT-backed $180^\circ$ loop and reports pixel and VLM consistency; and
\themebf{open-domain return}, which uses held-out edited first frames
and a compact $45^\circ$ leave/return probe.

\thinparagraph{Samples.}
The in-domain split provides first frames, metadata prompts, GT poses,
and GT images. This supports direct loop-closure PSNR and VLM checks for
fine-grained scene and identity details. The open-domain split has no
pixel-level GT trajectory, so we avoid treating pixel differences as
geometric truth and instead use controlled first-frame probes scored by
semantic verification.

\thinparagraph{Open-domain first-frame pool.}
We generate held-out first frames with Qwen Edit. For each of $20$
game-style environment prompts, the edit model inserts a distinctive
identity anchor while preserving the surrounding scene. Eight variants
per prompt yield a $20{\times}8$ pool whose objects are easy to name,
localize, and revisit.

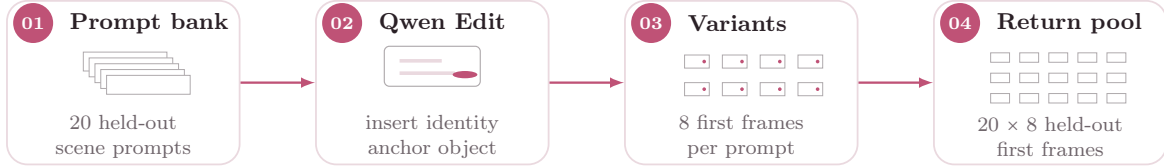
\begin{figure}[t]
\centering
\IfFileExists{fig/qwen_edit_first_frame_pipeline.tikz}{%
    \resizebox{0.96\linewidth}{!}{\definecolor{echoAccent}{HTML}{BF5276}
\definecolor{echoDark}{HTML}{2A2026}
\definecolor{echoMuted}{HTML}{6E646A}
\definecolor{echoLine}{HTML}{EAD9DF}

\begin{tikzpicture}[
  font=\small,
  box/.style={
    draw=echoLine,
    rounded corners=8pt,
    line width=0.7pt,
    fill=white,
    minimum width=3.2cm,
    minimum height=2.25cm,
    align=center,
    inner sep=8pt
  },
  badge/.style={
    circle,
    fill=echoAccent,
    text=white,
    font=\bfseries\scriptsize,
    minimum size=5.5mm,
    inner sep=0pt
  },
  title/.style={
    font=\bfseries\small,
    text=echoDark,
    align=center
  },
  note/.style={
    font=\footnotesize,
    text=echoMuted,
    align=center
  },
  flow/.style={
    -latex,
    draw=echoAccent,
    line width=0.9pt
  }
]

\node[box] (p) at (0,0) {};
\node[box] (e) at (4.25,0) {};
\node[box] (v) at (8.50,0) {};
\node[box] (s) at (12.75,0) {};

\foreach \n/\num/\name in {p/01/Prompt bank,e/02/Qwen Edit,v/03/Variants,s/04/Return pool} {
  \node[badge] at ([xshift=-1.24cm,yshift=0.82cm]\n.center) {\num};
  \node[title,anchor=west] at ([xshift=-0.86cm,yshift=0.82cm]\n.center) {\name};
}

\foreach \i in {0,...,4} {
  \draw[echoMuted!55,fill=white,line width=0.35pt]
    ([xshift={-0.55cm+0.08cm*\i},yshift={0.16cm-0.08cm*\i}]p.center)
    rectangle ++(1.15cm,0.26cm);
}
\node[note] at ([yshift=-0.74cm]p.center) {20 held-out\\scene prompts};

\draw[echoMuted!55,rounded corners=2pt,fill=white,line width=0.45pt]
  ([xshift=-0.66cm,yshift=-0.05cm]e.center) rectangle ++(1.32cm,0.55cm);
\draw[echoLine,line width=1.4pt] ([xshift=-0.45cm,yshift=0.30cm]e.center) -- ++(0.58cm,0);
\draw[echoLine,line width=1.4pt] ([xshift=-0.45cm,yshift=0.13cm]e.center) -- ++(0.76cm,0);
\fill[echoAccent] ([xshift=0.45cm,yshift=0.11cm]e.center) ellipse (0.17cm and 0.05cm);
\node[note] at ([yshift=-0.74cm]e.center) {insert identity\\anchor object};

\foreach \r in {0,1} {
  \foreach \c in {0,1,2,3} {
    \draw[echoMuted!55,fill=white,line width=0.35pt]
      ([xshift={-0.78cm+0.52cm*\c},yshift={0.20cm-0.38cm*\r}]v.center)
      rectangle ++(0.34cm,0.18cm);
    \fill[echoAccent] ([xshift={-0.51cm+0.52cm*\c},yshift={0.29cm-0.38cm*\r}]v.center)
      circle (0.025cm);
  }
}
\node[note] at ([yshift=-0.74cm]v.center) {8 first frames\\per prompt};

\foreach \r in {0,1,2} {
  \foreach \c in {0,1,2,3,4} {
    \draw[echoMuted!55,fill=white,line width=0.32pt]
      ([xshift={-0.82cm+0.40cm*\c},yshift={0.28cm-0.28cm*\r}]s.center)
      rectangle ++(0.26cm,0.13cm);
  }
}
\node[note] at ([yshift=-0.74cm]s.center) {20 $\times$ 8 held-out\\first frames};

\draw[flow] (p.east) -- (e.west);
\draw[flow] (e.east) -- (v.west);
\draw[flow] (v.east) -- (s.west);

\end{tikzpicture}}%
}{%
    \fbox{\parbox{0.92\linewidth}{\centering
    \textbf{[Figure placeholder: Qwen Edit first-frame pipeline]}\\[2pt]
    Held-out prompts are edited into identity-anchored first frames.}}%
}
\caption{\textbf{Qwen Edit construction of open-domain first frames.}
Each held-out environment prompt is edited into identity-anchored first
frames, which seed the open-domain return probe.}
\label{fig:qwen-edit-first-frame-pipeline}
\end{figure}

\thinparagraph{Metrics.}
Replay uses mean PSNR/SSIM/LPIPS over three generated chunks. In-domain
return matches mirrored frames along the outgoing and returning
trajectory and averages the pairwise image scores. Open-domain return
uses a visual-return PSNR proxy plus a Qwen3-VL-30B-A3B semantic score
for \textsc{Scene} and \textsc{Special Identity}. The VLM prompt weights
object appearance and presence more heavily than background plausibility:
\[
0.45\,s_\text{appearance}
+0.25\,s_\text{presence}
+0.20\,s_\text{view}
+0.10\,s_\text{scene},
\]
with the $[0,5]$ weighted sum rescaled to $[0,100]$. A cross-judge
sanity check is reported in \cref{sec:eval-vlm-robustness}.

\section{Additional Evaluation Details}
\label{sec:eval-appendix}

\begin{figure}[t]
\centering
\IfFileExists{fig/open_domain_revisit_panel.pdf}{%
    \includegraphics[width=\linewidth]{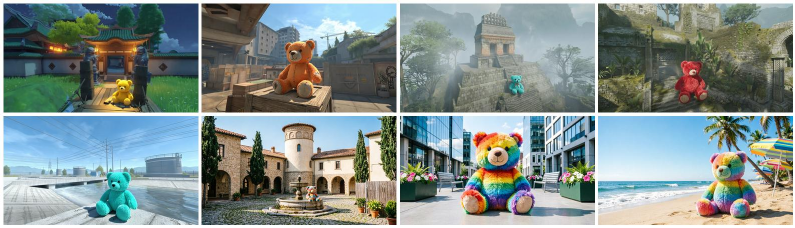}%
}{%
    \fbox{\parbox{0.92\linewidth}{\centering
    \textbf{[Figure placeholder: dense open-domain revisit panel]}\\[2pt]
    Representative first-frame sources from the open-domain pool.}}%
}
\caption{\textbf{Open-domain revisit source panel.} The $2{\times}4$
grid shows representative first-frame sources from the Qwen-edit
open-domain pool. The full pool is generated by editing distinctive
toy-like objects into held-out game-style environments; a practical
construction uses $20$ scene prompts and $8$ edited variants per prompt.
These objects serve as deliberately simple identity anchors for the VLM
judge: after the camera leaves and returns, the judge scores whether the
same object appearance, subject presence, camera viewpoint, and
background scene are preserved.}
\label{fig:open-domain-revisit-panel}
\end{figure}

\subsection{Robustness of the VLM Judge}
\label{sec:eval-vlm-robustness}

Because the open-domain branch relies on a single VLM judge, we run a
sanity check on whether the choice of judge changes our conclusions.
The probe is a constrained recognition task: the judge compares the
source frame with revisit-tail frames and decides whether the salient
object, viewpoint, and scene are preserved. We re-score a stratified
subset of open-domain cases with Qwen3-VL-30B-A3B, two stronger
closed-source judges, and a human anchor, all receiving the same images
and prompt. The alternative judges stay within a few points of Qwen3-VL
on average, remain correlated above $0.90$, and preserve the relative
reading of the subset. Changing the judge therefore adds at most a small
offset rather than reordering the memory profiles.

\begin{table}[t]
\centering
\small
\caption{\textbf{Cross-judge agreement for open-domain scoring.} Mean
scores over a shared open-domain subset. $\Delta$ is the mean absolute
difference from Qwen3-VL-30B-A3B, and $\rho$ is Pearson correlation
with Qwen3-VL.}
\label{tab:eval-vlm-cross-judge}
\begin{tabular}{lccc}
\toprule
\tableheader
\textbf{Judge} & \textbf{Mean score} & \textbf{$\Delta$ vs. Qwen} & \textbf{$\rho$ vs. Qwen} \\
\midrule
Qwen3-VL-30B-A3B & $19.6{\pm}13.9$ & --  & $1.00$ \\
Claude Opus 4.6  & $20.7{\pm}11.1$ & $3.1$ & $0.93$ \\
GPT-5.5          & $21.1{\pm}11.4$ & $2.7$ & $0.94$ \\
Human            & $20.9{\pm}11.9$ & $1.3$ & $0.96$ \\
\bottomrule
\end{tabular}
\end{table}

\section{Results}
\label{sec:main-results}

\newcommand{\tabmainbest}[1]{{\bfseries #1}}

The experiments are organized around the failure mode that motivates the
paper: a model can generate plausible local video while still failing to
remember the world at revisit time. Each table reports the same evidence
bundle: Replay PSNR/SSIM/LPIPS (R-P/R-S/R-L), in-domain return
PSNR/SSIM/LPIPS (ID-P/ID-S/ID-L), and open-domain VLM score (O-V).
The first two branches verify camera-following and loop closure; the
last branch is the strongest semantic memory stress test.

\subsection{Memory Families}
\label{sec:main-family}

\begin{table}[t]
\centering
\scriptsize
\caption{\textbf{Representative memory-family comparison.} R-P/R-S/R-L
denote Replay PSNR/SSIM/LPIPS; ID-P/ID-S/ID-L denote in-domain closure
PSNR/SSIM/LPIPS; O-V denotes open-domain VLM.}
\label{tab:main-headline}
\resizebox{\linewidth}{!}{%
\begin{tabular}{l|ccc|ccc|c}
\toprule
\tableheader
 & R-P$\uparrow$ & R-S$\uparrow$ & R-L$\downarrow$ & ID-P$\uparrow$ & ID-S$\uparrow$ & ID-L$\downarrow$ & O-V$\uparrow$ \\
\midrule
I2V baseline & 10.03 & 0.398 & 0.534 & 10.32 & 0.291 & 0.643 & 12.25 \\
Spatial Memory baseline      & \tabmainbest{13.60} & 0.411 & 0.554 & 10.04 & 0.260 & 0.617 & 6.00 \\
Compression weight-only      & 8.65 & 0.174 & 0.752 & 10.67 & 0.154 & 0.645 & 22.38 \\
State-Space (legacy hybrid)  & 12.69 & 0.344 & 0.581 & \tabmainbest{12.23} & 0.298 & 0.584 & 34.75 \\
State-Space (block-wise)     & 9.59 & 0.282 & 0.698 & 11.95 & 0.280 & \tabmainbest{0.535} & \tabmainbest{69.00} \\
Context learning, $K{=}5$    & 11.92 & 0.408 & 0.501 & 10.72 & 0.307 & 0.596 & 50.75 \\
Context learning, $K{=}20$   & 12.54 & \tabmainbest{0.449} & \tabmainbest{0.496} & 11.07 & \tabmainbest{0.359} & 0.543 & 58.63 \\
\bottomrule
\end{tabular}%
}
\end{table}

The headline comparison shows why replay alone is insufficient.
Different columns select different winners: Spatial Memory has the best
Replay PSNR, long raw Context has the best Replay SSIM/LPIPS, and
block-wise State-Space has the strongest open-domain VLM score. This
inversion means that memory quality is not the same thing as image
quality under GT motion. Replay tells us whether the generator can
follow an observed trajectory; open-domain return tells us whether it
can put the same object and scene back after leaving the view.

This disagreement is not noise in the table; it is the central
diagnostic signal. Replay rewards the model for matching a known camera
trajectory and reconstructing visible regions with low distortion. A
return probe removes part of that support. The model must leave the
relevant evidence behind, generate intervening frames, and then
reconstruct the same world state when the viewpoint comes back. The
Spatial--State-Space inversion is therefore informative: Spatial Memory
can improve local reconstruction, but the open-domain probe asks it to
recover a distinctive object from a compact scene summary. Block-wise
State-Space produces less faithful pixels under GT replay, but its
recurrent state appears to preserve the semantic identity needed for
return.

The Context rows are the key capacity baseline. Moving from the
anchor-only I2V baseline to $K{=}5$ raises open-domain VLM from $12.25$
to $50.75$, while $K{=}20$ reaches $58.63$. The replay image bundle
improves much less dramatically. This asymmetry suggests that raw
history mainly helps semantic return rather than merely improving local
pixel fidelity. Compact memories should therefore be compared against
raw Context, not only against I2V.

\begin{figure}[t]
\centering
\IfFileExists{fig/replay_revisit_metric_alignment.pdf}{%
    \includegraphics[width=\linewidth]{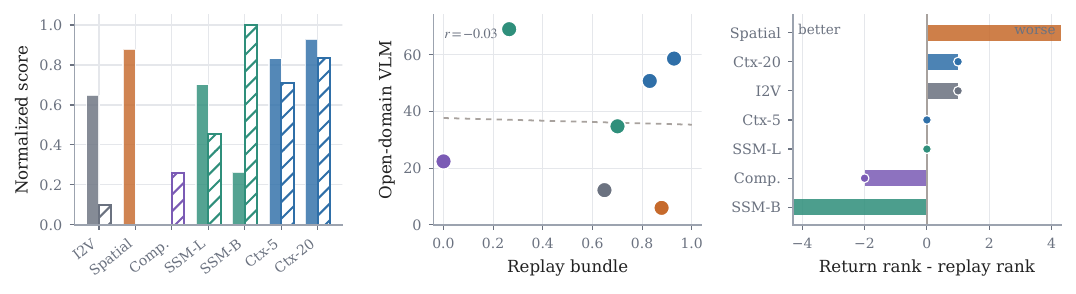}%
}{%
    \fbox{\parbox{0.92\linewidth}{\centering
    \textbf{[Figure placeholder: replay--return alignment]}}}%
}
\caption{\textbf{Replay is a health signal, not the final memory
score.} The normalized replay and return views disagree across memory
families, showing that low-level trajectory fidelity does not determine
semantic revisit consistency.}
\label{fig:main-replay-return-alignment}
\end{figure}

The alignment plot turns the headline table into a diagnostic rather
than a ranking. Replay catches broken generators that cannot follow a
camera trajectory, so it remains a useful and inexpensive health signal.
However, the rank shift from replay to return shows why it cannot be
used as the final memory score: the strongest replay rows can still
forget the salient object, while rows with weaker replay fidelity can
carry scene evidence through the excursion.

\finding{1}{\emph{Open-domain return is the most discriminative memory
stress test.} Replay and in-domain return are necessary filters, but the
open-domain VLM column is where memory families separate most clearly.}

\begin{figure}[t]
\centering
\IfFileExists{fig/representative_sweep_panel.pdf}{%
    \includegraphics[width=\linewidth]{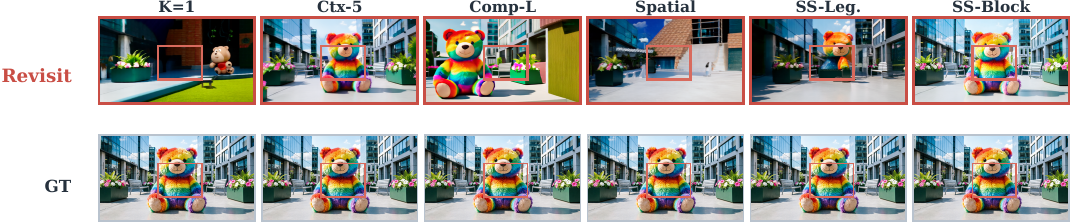}%
}{%
    \fbox{\parbox{0.92\linewidth}{\centering
    \textbf{[Figure placeholder: representative sweep panel]}}}%
}
\caption{\textbf{Representative open-domain revisit case.} Matched
revisit frames expose whether the salient object survives the
leave-and-return motion. The qualitative panel illustrates why high
Replay fidelity need not imply high semantic return.}
\label{fig:main-representative-sweep}
\end{figure}

\subsection{Spatial Read-out}
\label{sec:main-spatial}

\begin{table}[t]
\centering
\scriptsize
\caption{\textbf{Spatial Memory read-out ablation.}}
\label{tab:main-spatial}
\resizebox{\linewidth}{!}{%
\begin{tabular}{llccc|ccc|c}
\toprule
\tableheader
 & Read-out & R-P$\uparrow$ & R-S$\uparrow$ & R-L$\downarrow$ & ID-P$\uparrow$ & ID-S$\uparrow$ & ID-L$\downarrow$ & O-V$\uparrow$ \\
\midrule
Spatial baseline        & default & 13.60 & 0.411 & 0.554 & 10.04 & 0.260 & 0.617 & 6.00 \\
Spatial inject-none     & none & \tabmainbest{14.66} & \tabmainbest{0.417} & \tabmainbest{0.541} & \tabmainbest{10.38} & \tabmainbest{0.309} & 0.631 & 15.50 \\
Spatial concat-text     & text KV concat & 13.44 & 0.412 & 0.572 & 9.35 & 0.272 & \tabmainbest{0.614} & 10.25 \\
Spatial cross-attn RO   & separate cross-attn & 13.52 & 0.409 & 0.557 & 8.49 & 0.232 & 0.664 & \tabmainbest{17.12} \\
\bottomrule
\end{tabular}%
}
\end{table}

Spatial summaries are efficient, but the ablation shows that efficiency
does not imply semantic memory. The inject-none row obtains the best
Replay bundle despite withholding the stored tokens from the generator,
which indicates that replay improvements can come from regularization
rather than usable memory. Dedicated cross-attention improves
open-domain VLM to $17.12$, but the family remains far below raw
Context and block-wise State-Space. The bottleneck is therefore shared
between storage bandwidth and read-out access: a compact grid can encode
where things are, but it does not reliably encode which object must be
regenerated after return.

This distinction is important because spatial memories are often
attractive precisely when long raw context is too expensive. The replay
numbers make them look promising: the family can reconstruct local
trajectory frames well, and the inject-none row even improves the
low-level image bundle. The return probe changes the interpretation.
When the salient object must be regenerated from memory, a low-bandwidth
scene grid under-specifies identity. The result is not that spatial
memory is useless; it is that a scene summary needs object-aware
retention and an explicit read-out path before it can replace raw visual
evidence.

\finding{2}{\emph{Spatial summaries are efficient but not yet reliable
as semantic memory.} The open-domain gap points jointly to storage
bandwidth and read-out design; replay efficiency alone cannot identify
which factor is binding.}

The practical implication is that future spatial memories should be
tested at two interfaces. First, the write path should be asked whether
it retains object-level evidence rather than only coarse layout. Second,
the read path should be asked whether the generator can select and use
the relevant token at return time. A single replay score mixes these
questions together. In our matrix, the dedicated read-out row improves
the semantic column but does not close the gap, which suggests that both
sides need to change: the storage grid should be more identity-aware,
and the read-out should expose object evidence in a form the diffusion
transformer can directly condition on.

\subsection{Compression}
\label{sec:main-compression}

\begin{table}[t]
\centering
\scriptsize
\caption{\textbf{Compression ablation.}}
\label{tab:main-compression}
\resizebox{\linewidth}{!}{%
\begin{tabular}{llccc|ccc|c}
\toprule
\tableheader
 & Mechanism & R-P$\uparrow$ & R-S$\uparrow$ & R-L$\downarrow$ & ID-P$\uparrow$ & ID-S$\uparrow$ & ID-L$\downarrow$ & O-V$\uparrow$ \\
\midrule
Weight-only baseline & token weighting & 8.65 & 0.174 & 0.752 & \tabmainbest{10.67} & 0.154 & 0.645 & 22.38 \\
Length $r{=}2$       & temporal length & 7.84 & 0.162 & 0.746 & 8.82 & 0.183 & 0.654 & 24.00 \\
Length $r{=}4$       & temporal length & \tabmainbest{9.88} & 0.183 & \tabmainbest{0.714} & 9.53 & 0.163 & \tabmainbest{0.617} & \tabmainbest{43.25} \\
Hybrid $r{=}2$ + weight & both & 9.10 & 0.181 & 0.734 & 7.87 & \tabmainbest{0.277} & 0.695 & 8.75 \\
Hybrid $r{=}4$ + weight & both & 9.63 & \tabmainbest{0.215} & 0.730 & 8.81 & 0.156 & 0.625 & 9.00 \\
\bottomrule
\end{tabular}%
}
\end{table}

Compression is not a monotone capacity dial. Weight-only compression
preserves every temporal position, yet length compression with
$r{=}4$ performs much better on open-domain VLM. A shorter but cleaner
context can be easier for the generator to read than a full temporal
stack with diffuse learned weights. The hybrid rows are the clearest
failure case: pooling removes temporal evidence before weighting can
decide which views matter, so the returned object is often
under-specified. This result argues against treating all compression
operators as interchangeable memory-saving tricks.

The compression table also explains why ``same budget'' comparisons are
not enough. A method can preserve many tokens but still spread attention
over frames that are only weakly useful for the return. Conversely, a
more aggressive temporal reduction can keep fewer but cleaner slots,
which makes the evidence easier for the generator to consume. The hybrid
rows show a negative interaction between two individually plausible
operators: first reducing temporal support, then learning weights over
the reduced sequence, can erase the very view that should have anchored
the return. A useful compressed memory therefore needs an objective for
\emph{what} to retain, not only a smaller token count.

\finding{3}{\emph{Compression must preserve return-critical evidence,
not just reduce cost.} Length compression can beat token weighting, but
hybrid compression shows that stacking budget-saving operators can
destroy the identity evidence needed for revisit.}

\subsection{Implicit and Raw Memory}
\label{sec:main-implicit-context}

\begin{table}[t]
\centering
\scriptsize
\caption{\textbf{State-Space and raw-context ablations.}}
\label{tab:main-implicit-context}
\resizebox{\linewidth}{!}{%
\begin{tabular}{llccc|ccc|c}
\toprule
\tableheader
 & Setting & R-P$\uparrow$ & R-S$\uparrow$ & R-L$\downarrow$ & ID-P$\uparrow$ & ID-S$\uparrow$ & ID-L$\downarrow$ & O-V$\uparrow$ \\
\midrule
State-Space & legacy hybrid & 12.69 & 0.344 & 0.581 & \tabmainbest{12.23} & 0.298 & 0.584 & 34.75 \\
State-Space & block-wise & 9.59 & 0.282 & 0.698 & 11.95 & 0.280 & \tabmainbest{0.535} & \tabmainbest{69.00} \\
\midrule
Context & $K{=}1$ & 10.03 & 0.398 & 0.534 & 10.32 & 0.291 & 0.643 & 12.25 \\
Context & $K{=}5$ & 11.92 & 0.408 & 0.501 & 10.72 & 0.307 & 0.596 & 50.75 \\
Context & $K{=}20$ & 12.54 & \tabmainbest{0.449} & \tabmainbest{0.496} & 11.07 & \tabmainbest{0.359} & 0.543 & 58.63 \\
\bottomrule
\end{tabular}%
}
\end{table}

The State-Space comparison is the largest controlled open-domain jump
in the matrix. The legacy hybrid row has better replay PSNR and strong
in-domain PSNR, but block-wise recurrence nearly doubles open-domain
VLM. This suggests that implicit memory is not a single design choice:
the state must be structurally integrated enough that the network cannot
bypass it when the camera leaves visible support.

The replay cost of block-wise recurrence is also informative. A
non-bypassable state competes with local reconstruction capacity: the
network must summarize history, update that summary, and use it during
generation. This can hurt pixel metrics on GT trajectories, where the
best strategy is often to focus on local visual fidelity. The same
structural commitment becomes valuable in open-domain return, where the
model has no pixel-aligned target and must rely on carried evidence.
This trade-off is exactly why a single scalar would be misleading. A
design that looks worse under replay can be better at the task that
motivated memory in the first place.

The raw-context rows calibrate the cost of compactness. Most of the
open-domain gain appears by $K{=}5$, and $K{=}20$ adds further
improvement while giving the best replay SSIM/LPIPS. This is a
conservative but demanding baseline: specialized memory mechanisms
should be judged by how much of the $K{=}20$ return benefit they
retain. Under that bar, only block-wise State-Space exceeds raw Context
on open-domain VLM.

The context curve also prevents an overly pessimistic reading of compact
memory. Raw context is strong, but the gain is sublinear: most of the
open-domain improvement arrives by $K{=}5$, with $K{=}20$ adding a
smaller but still meaningful margin. This shape suggests that there is
room for compact memories to match raw context if they retain the right
evidence. The current matrix does not show that compression or spatial
summaries are hopeless; it shows that generic compression and generic
scene grids are not yet the right abstractions for return. The target is
now clearer: preserve high-value views and object anchors while avoiding
the linear cost of long raw context.

\finding{4}{\emph{Block-wise recurrence is the strongest current
open-domain memory bias.} The gain over legacy recurrence shows that
the structure of implicit memory matters, not merely the decision to
make memory recurrent.}

\finding{5}{\emph{Raw context is the capacity baseline, and the gain it
buys is overwhelmingly on revisit rather than on Replay.} Compact and
implicit memories should be judged by how much of the $K{=}20$
open-domain benefit they retain, not by how much they beat I2V.}

\subsection{Scaling and Efficiency}
\label{sec:main-scaling-efficiency}

\begin{figure}[t]
\centering
\begin{minipage}[t]{0.49\linewidth}
\centering
\IfFileExists{fig/context_length_scaling.pdf}{%
    \includegraphics[width=\linewidth]{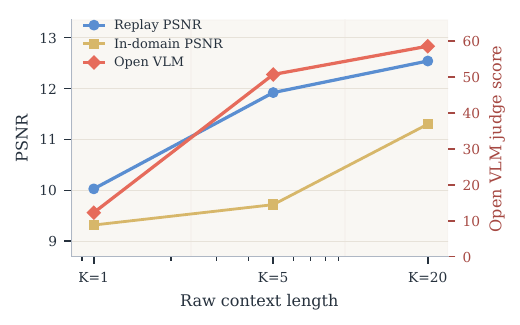}%
}{%
    \fbox{\parbox{0.92\linewidth}{\centering
    \textbf{[Figure placeholder: context length scaling]}}}%
}
\end{minipage}\hfill
\begin{minipage}[t]{0.49\linewidth}
\centering
\IfFileExists{fig/training_time_efficiency.pdf}{%
    \includegraphics[width=\linewidth]{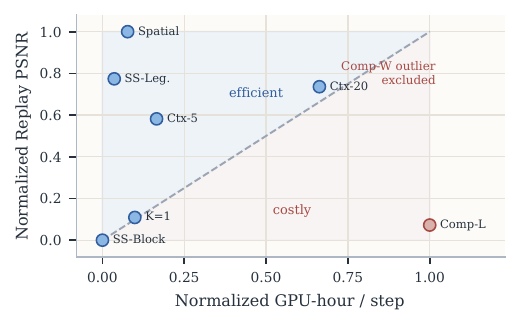}%
}{%
    \fbox{\parbox{0.92\linewidth}{\centering
    \textbf{[Figure placeholder: training-time efficiency]}}}%
}
\end{minipage}
\caption{\textbf{Context scaling and training efficiency.} Left:
increasing raw context improves open-domain semantic return much more
than the replay image bundle. Right: normalized replay PSNR versus
normalized GPU-hour per step illustrates that replay efficiency and
semantic memory are different optimization targets.}
\label{fig:main-context-efficiency}
\end{figure}

The context-scaling curve makes the capacity result more explicit. The
open-domain VLM jump from $K{=}1$ to $K{=}5$ is much larger than the
corresponding replay gain, and $K{=}20$ continues to improve return
while only modestly improving low-level trajectory metrics. This slope
mismatch is one of the cleanest signs that revisit consistency is a
separate capability. If the problem were only local reconstruction, the
replay and open-domain curves would move together. Instead, raw evidence
primarily helps the model restore the same object and scene after the
camera leaves.

The efficiency panel prevents the opposite over-correction. Raw context
is a strong baseline, but it is not free: increasing $K$ raises memory
and training cost. Spatial summaries and compression remain worth
studying because they target this cost, even though the current
instantiations do not yet preserve enough semantic evidence. Block-wise
State-Space occupies a different point in the trade-off: it is not the
most replay-efficient row, but it is the only compact or implicit
mechanism that exceeds raw Context $K{=}20$ on open-domain VLM. The
practical goal is therefore not simply to maximize replay quality per
GPU-hour, but to find mechanisms that preserve return-critical evidence
at a cost below raw context.

This trade-off matters for how the matrix should be used. A researcher
who only wants short GT-trajectory replay might choose a different row
from a researcher who wants revisitable open-domain scenes. Spatial
Memory is attractive in a replay-efficiency view because it provides
strong local reconstruction with a compact representation. Context
$K{=}20$ is attractive as a robustness baseline because it exposes the
generator to many uncompressed observations. Block-wise State-Space is
attractive as a memory mechanism because it carries scene evidence
without appending a long explicit history. None of these rows dominates
all axes, and that is the point: memory design is a multi-objective
problem involving local fidelity, revisit consistency, and cost.

The current results also suggest how to design the next controlled
matrix. First, compressed methods should be optimized against a
return-aware signal, not only against replay loss or token count.
Second, spatial methods should report object-level retention and
read-out access separately. Third, recurrent methods should vary how
deeply the state is integrated into the DiT rather than treating
``state-space memory'' as one monolithic option. Finally, every compact
method should be compared against a raw-context scaling curve. Without
that curve, it is too easy to mistake a win over I2V for a win over a
strong memory baseline.

\subsection{Cross-cutting Observations}
\label{sec:main-cross-cutting}

Three patterns are consistent across the ablations. First, low-level
image quality is necessary but not sufficient. Rows that fail Replay are
not useful world models, but rows that win Replay can still forget the
scene. Second, memory storage and memory read-out should be evaluated
separately. The Spatial rows show that storing a compact representation
does not guarantee that the generator can use it at the return point.
Third, capacity baselines should be strong. The I2V baseline is useful
as a floor, but the real reference for compact memories is raw Context
$K{=}20$, because it shows what the same generator can do when given
uncompressed evidence under the same retrieval and action interface.

These observations explain why the paper reports a matrix rather than a
single winning model. The rows expose different failure modes:
Spatial Memory preserves local replay structure but loses identity;
Compression can make history cheaper but may delete the decisive view;
legacy recurrence smooths trajectories without carrying enough
open-domain evidence; block-wise recurrence carries that evidence but
pays a replay-quality cost; and raw Context remains a stubborn capacity
baseline. A memory mechanism for action world models should therefore
be judged by how it moves along all three axes: trajectory following,
in-domain return, and open-domain semantic return.

\subsection{Design Implications}
\label{sec:main-design-implications}

The controlled matrix suggests a concrete checklist for future memory
modules. A new memory mechanism should first be compared against raw
Context at multiple $K$ values. This avoids the common but weak claim
that a method improves over anchor-only I2V. The stronger question is
whether the method retains the semantic return benefit of raw context
while using fewer tokens, less compute, or a more stable state. Second,
the mechanism should report write capacity and read-out capacity
separately. Spatial Memory shows that a model can store a compact scene
summary without making the right evidence available to the generator at
return time. Third, compression should be evaluated by what evidence it
preserves, not only by how many tokens it removes. A good compressor for
world models is not merely a smaller context; it is a selective memory
that keeps views and objects likely to become load-bearing later.

The results also argue for training signals that are closer to revisit
quality. The current single-stage objective supervises target-frame
denoising, while revisit consistency is measured after generation. This
gap explains why replay can improve without semantic return improving.
One natural direction is to add return-aware auxiliary supervision:
object-level retention losses for spatial summaries, contrastive
alignment between source and return frames, or VLM-guided selection
signals for compression. Another direction is to expose recurrent states
to explicit object or view prediction during training, so that the state
is pressured to encode the evidence that will be needed after the
camera leaves visible support.

Finally, the evaluation should remain multi-branch even when stronger
models arrive. Better video backbones will raise replay quality and may
make longer open-domain loops feasible, but they will not remove the
distinction between following and remembering. If anything, stronger
generators make the distinction more important: a highly capable model
can hallucinate plausible replacements for missing objects, making
qualitative failures harder to notice without a return-specific
semantic check. The three-branch protocol is therefore not only a
diagnostic for the present matrix; it is a guardrail for future systems
that may look visually convincing while still failing to preserve the
same world.

\section{Discussion}
\label{sec:conclusion}

We have presented \method{}, a controlled study of memory in action
world models. By fixing the backbone, optimizer, sampler, and
evaluation protocol, the study makes the memory/context profile the
main experimental variable. The resulting matrix separates a model's
ability to follow camera actions from its ability to preserve the
world that those actions leave behind.

\thinparagraph{Main findings.}
Three lessons stand out. First, raw context is a strong capacity
baseline: increasing context improves open-domain revisit more reliably
than it improves replay metrics. Second, compactness alone is not enough:
Spatial Memory and hybrid compression are efficient, but lose identity
evidence without object-aware retention and read-out. Third, implicit
memory depends on structure; block-wise recurrence preserves open-domain
return much better than lightweight recurrent smoothing.

\thinparagraph{Evaluation lesson.}
Replay quality and revisit quality are not monotonically aligned.
Several rows that look competitive under low-level replay metrics
fall behind under semantic revisit scores. Selecting models by Replay
alone would therefore favor the wrong family for interactive,
revisitable world generation. This is why we report replay,
in-domain return, and open-domain return as separate evidence rather
than collapsing them into one scalar.

\thinparagraph{Limitations.}
The study is still bounded in scope. We train on a single dataset,
so the ranking of mechanisms may change with noisier poses, different
camera statistics, larger compute, or multi-stage curricula. The
open-domain scores also rely on a VLM judge; our cross-judge check
(\cref{sec:eval-vlm-robustness}) is encouraging, but broader human
calibration would make the comparison stronger. Finally, revisit
quality is not yet available as a cheap training-time signal, so
model selection still has to lean on imperfect proxies. The reported
rows should therefore be read as representative points in a controlled
matrix, not as the final implementations of each memory family.

\thinparagraph{Outlook.}
The next step is to make memory design explicitly revisit-aware:
compression should preserve high-value object evidence, spatial
summaries should expose object-level read-out to the generator, and
structured recurrence should be evaluated as a first-class design
axis. More broadly, \method{} argues that world models should be measured
not only by how smoothly they continue a trajectory, but by whether they
can return to the same world after leaving it.

\clearpage
\bibliographystyle{plainnat}
\bibliography{cite}

\clearpage
\appendix
\section{Action World Models: Preliminaries and Related Work}
\label{sec:related}

The key components of an action world model are the
\emph{video diffusion backbone}, the \emph{action representation}
and its injection site, the \emph{memory subsystem}, the
\emph{retrieval policy}, the \emph{training schedule}, and the
\emph{evaluation and benchmarking protocol}. Each component has its
own intricacies, and the interactions among them are non-trivial. Our
study investigates them under a single backbone and a single training
budget, with the explicit goal of isolating the contribution of memory.

World models have been studied as latent imagination, predictive
state dynamics, and joint-embedding prediction~\cite{ha2018world,lecun2022path,hafner2023mastering}.
Recent video and embodied systems extend this view from compact latent
states to generated visual rollouts conditioned on prompts, camera
motion, or agent actions~\cite{zhu2024sora,hu2025simulating,yu2025survey,li2025comprehensive,long2025survey,feng2025survey}.
This shift makes memory a practical modeling question: a rollout must
not only synthesize the next plausible clip, but also preserve scene and
object evidence after the camera leaves and later revisits a region.

\thinparagraph{Diffusion-Transformer Video Backbones.}
A modern video generator typically follows the latent-diffusion
recipe: an autoencoder maps each frame to
a low-dimensional latent, and a denoising network learns the
conditional velocity field over a stack of latent frames. Recent
work has converged on diffusion-transformer
backbones~\cite{wan2025wan,alibaba2025wan25}, which scale
favorably and natively expose per-frame token positions. Compared
to U-Net-based 3D diffusion models, DiT backbones are particularly
well suited to action-controlled generation: any per-frame signal
(camera RT, action labels) can be attached to the corresponding
frame token and propagated through self-attention, and the choice
of injection site directly
determines how strongly the signal reaches the spatial attention.
Throughout this paper we use an off-the-shelf video
DiT~\cite{wan2025wan} as the common substrate; all design-space
choices we study are local edits to this substrate. The broader
video-generation ecosystem has converged on increasingly powerful
diffusion-transformer backbones, including
Imagen Video~\cite{ho2022imagen}, Sora~\cite{openai2024sora,openai2025sora2},
Veo~\cite{google2025veo3}, Wan~\cite{wan2025wan,alibaba2025wan25},
Seedance~\cite{gao2025seedance}, Waver~\cite{zhang2025waver},
joint audio-video DiT~\cite{liu2025javisdit}, and camera-controllable
generative re-rendering~\cite{bai2025recammaster,luo2025camclonemaster}.
Native multimodal models such as Emu3.5~\cite{cui2025emu3} explore
unified token-based modeling across images, video, and language. Cosmos
is better described as a world-foundation-model platform for Physical
AI, with video foundation models for prediction and conditional world
generation rather than a single native multimodal model~\cite{agarwal2025cosmos,ali2025world}.

\thinparagraph{Action and Camera Conditioning.}
A second axis, orthogonal to the memory subsystem, is the
representation of action signals. Camera-controllable video
diffusion has explored a spectrum of representations: world-frame
extrinsics in context-based action-world models~\cite{yu2025context}, Pl\"ucker
embeddings, dense optical flow,
and discretized action tokens for game-style control as in
game-style world models. Time-aware positional
encoding~\cite{xu2026ucm}, cube-map and panoramic
parametrizations~\cite{li2026cubecomposer}, and iterative
video editing on the same scene~\cite{lee2026memory}
further illustrate alternative action and viewpoint
representations. We adopt the most
straightforward representation, $12$-dimensional relative-RT camera
poses, and treat the \emph{injection site} as a first-class design
variable: post-attention, pre-norm, pre-QKV, or gated
pre-QKV (\cref{sec:method-design-space}). This mirrors a finding from
the multimodal-LLM literature: a seemingly small
plumbing decision (\eg, where exactly the connector reads from)
often dominates large architectural ones.

\thinparagraph{Memory for Long-Horizon Video Generation.}
Long-horizon video generation has increasingly treated memory as an
interface between the current denoising window and observations that are
no longer visible. One direct strategy is retrieval: previous frames or
features are selected by camera overlap, entity identity, temporal
proximity, or scene geometry, and then appended as context for the next
chunk~\cite{yu2025context,hong2025relic,lee20253d,zhao2025spatia,hu2026geometry}.
This line has been extended to streaming and interactive settings, where
the system must decide what to keep in a fast cache, what to compress,
and when to refresh evidence during rollout~\cite{wu2026infinite,sun2025worldplay,huang2025memoryforcing,chen2025learningwm}.
Other work reduces the cost of carrying history by weighting tokens,
packing temporal windows, pooling latent frames, or replacing a full
frame stack with compact spatial summaries~\cite{cai2025mixture,wu2025pack,ji2025memflow,wu2025video}.
Keyframe memories and multi-shot generation systems make a similar
trade-off at the level of scenes or shots rather than individual
tokens~\cite{meng2025holocine,an2025onestory,zhang2025storymem}.
These methods differ in implementation, but they share the same
underlying trade-off: the model must preserve enough visual evidence for
a later return while keeping the current generation window affordable.

A complementary direction avoids explicit retrieval at inference time
and carries history in a recurrent or learned state. State-space video
world models propagate a causal hidden state through selected DiT blocks,
whereas recurrent gates, test-time fast weights, flow-equivariant states,
and plug-and-play memory tokens offer lighter or more modular
variants~\cite{yu2025videossm,chen2025recurrent,zhang2025test,lillemark2026flow,song2025learning}.
Persistent-scene editors and spatial world memories similarly emphasize
that what matters is not merely whether information is stored, but
whether the generator can read it back when the viewpoint returns~\cite{po2026multigen,savva2026solaris,inspatio2026inspatio,chen2026teleworld,fu2026plenoptic}.
We use this literature to define a controlled comparison rather than a
new memory module: all rows share the same backbone, camera interface,
training budget, sampler, and evaluation protocol, while the memory
representation and read-out path are varied.

\thinparagraph{Evaluation and Benchmarking.}
Static image and single-clip video metrics such as PSNR, SSIM, LPIPS,
FID, and FVD remain the lingua franca of video
generation, but on their own they do not capture long-horizon
\emph{revisit consistency}. Context-based action-world-model work has begun reporting
specialized revisit metrics on rotation
loops~\cite{yu2025context,hong2025relic,huang2025memoryforcing,sun2025worldplay}, while the broader vision-language
community has shifted toward \emph{model-as-judge} protocols using
strong VLMs (\eg, GPT-4V or
Qwen3-VL-30B-A3B~\cite{xu2025qwen3,wang2024qwen2}). Multi-shot storytelling
benchmarks~\cite{meng2025holocine,zhang2025storymem,an2025onestory,zhou2026videomemory},
streaming long-video efficiency benchmarks~\cite{yang2025longlive,zhu2025memorize,zhen2026soulx},
and out-of-sight dynamics evaluation~\cite{duan2026liveworld,chen2026out}
further enrich this evaluation landscape. Spatial reasoning benchmarks
and general VLM/LLM backbones also inform model-as-judge protocols for
semantic verification~\cite{ma2025spatialreasoner,chen2024spatialvlm,deitke2025molmo,xu2025qwen3,comanici2025gemini,openai2025gpt5}. Our evaluation
(\cref{sec:evaluation}) combines both: long-horizon GT replay for
PSNR/SSIM/LPIPS image quality, GT-backed in-domain $180^\circ$ loop closure
with revisit PSNR and VLM scoring, and open-domain $45^\circ$
return probes measured by revisit PSNR proxies and VLM scores for
\textsc{Scene} and \textsc{Special Identity} consistency.

\thinparagraph{How \method{} differs.}
Our contribution is neither a new storage mechanism, nor a new
retriever, nor a new conditioning trick. It consists of (i) a
\emph{common interface} into which all of the above can be plugged
without modifying the training loop, sampler, or evaluation protocol;
(ii) a \emph{matched ablation protocol} that exercises the full design
space under a single-stage training budget; and (iii) a
\emph{three-branch evaluation stack} that links trajectory replay,
in-domain revisit, and open-domain revisit consistency. Rather than
introducing yet another isolated model, the study is intended to make
the interactions among storage, compression, read-out, recurrence, and
evaluation visible within a single experimental matrix.

\section{Training and In-training Diagnostics}
\label{sec:training}

By fixing the backbone, optimizer, schedule, and inference path across
variants (\cref{sec:method-registry}), the training setup is a
\emph{constant} of the study: the only thing that varies across the
rows of \cref{tab:abl-headline} is the memory module. This is what
lets the later differences be read as storage, compression, read-out,
or recurrence effects rather than as hidden protocol changes. We
describe that constant below, and then promote in-training replay
sampling from a debugging convenience to a diagnostic signal.

\subsection{Data}
\label{sec:training-data}

We train on the Context-as-Memory dataset~\cite{yu2025context}, which
contains real-world camera trajectories paired with per-frame poses
and textual prompts. Each training sample is an $81$-frame segment
at $352{\times}640$ resolution, drawn from a long source video. For
each sample, the data interface provides
(i) the target segment $\mathbf{x}_{s:e}$,
(ii) the corresponding $12$-dim relative-RT camera trajectory, and
(iii) a context buffer of $K{-}1$ historical frames retrieved from
pre-computed field-of-view overlap labels keyed on the mid-frame of
the segment.

\thinparagraph{Train/eval reference-frame alignment.}
A non-trivial implementation invariant is that the same notion of
\emph{reference frame} is used at training and evaluation time when
retrieving context: training uses $r_\text{train} = (s+e)/2$, while
in-domain evaluation uses $r_\text{eval} = N_\text{frames}/2$ on
videos that always start at frame $0$. This alignment is not a
headline ablation, but we still treat it as a hard protocol
invariant.

\subsection{Single-stage Optimization}
\label{sec:training-optim}

All variants in this study are trained in a single stage, of
identical length, with the optimizer settings listed in
Table~\ref{tab:hparams}. We deliberately avoid multi-stage curricula
and stage-1 / stage-2 splits: in pilot experiments at this scale, we
were unable to find a multi-stage schedule that consistently
outperformed the single-stage one, and the simpler protocol makes
the ablations easier to read.

\begin{table}[t]
\centering
\small
\caption{Single-stage hyper-parameters shared by all variants. Only
the columns marked \emph{(varies)} are perturbed across variants.}
\label{tab:hparams}
\begin{tabular}{ll}
\toprule
\tableheader
\textbf{Setting} & \textbf{Value} \\
\midrule
Backbone                                & Video DiT (per-frame VAE) \\
Resolution                              & $352\times 640$ \\
Segment length $T$                      & $81$ frames \\
Context length $K$                      & \emph{(varies)} -- $\{1, 5, 20\}$, default $5$ \\
Memory module                           & \emph{(varies)} -- Context, Compression, Spatial, or State-Space \\
Optimizer                               & AdamW \\
Learning rate                           & $5\!\times\!10^{-5}$ ($1\!\times\!10^{-4}$ at $K{=}1$) \\
Per-device batch / grad accumulation    & $1$ / $1$ \\
GPUs                                    & $8$ (A100-80G) \\
Total steps                             & $5\,$k \\
Timestep shift                          & $15$ \\
Spike rejection threshold               & $15.0$ \\
Target-frame-only supervision           & enabled \\
Flow noise shift                         & $1.0$ \\
Relative-RT action encoding              & enabled \\
Context placement                        & suffix \\
$10\%$ overlap-drop policy              & enabled \\
\bottomrule
\end{tabular}
\end{table}

\subsection{Replay Sampling Methodology}
\label{sec:sampling-progress}

A practical observation from pilot experiments is that the scalar
flow-matching loss is a poor progress indicator for action-world-model
training. Two runs with visually different boundary behavior can have
nearly indistinguishable denoising-loss curves; conversely, a run with
a misconfigured camera-action condition can look stable on the loss
until a return trajectory exposes the error. We therefore use replay
sampling as an in-training methodology, not as an informal debugging
video. The replay protocol fixes the sample pool, the action
construction, and the generation code path, so changes in the sampled
videos can be attributed to the memory profile under training, and
boundary continuity, identity preservation, and action--visual
alignment become directly comparable across variants well before final
evaluation.

\thinparagraph{Sample pool and segment length.}
Every variant in Sec.~\ref{sec:ablation} is configured to sample, at
fixed step intervals, videos built from the same held-out first
frames. Each generated segment contains $81$ frames, matching the
training target length, and multi-segment replay concatenates several
such segments without changing the memory interface between segments.
This keeps replay close to the actual inference regime, rather than
measuring a short single-clip proxy.

\thinparagraph{Camera-trajectory construction.}
Replay actions are produced by the same relative-RT constructor used
in evaluation. For ground-truth replay, the constructor reads the
dataset camera trajectory and converts each frame into a $12$-dim
relative RT vector. For return probes, it synthesizes symmetric yaw
motions, including $180^\circ$/$360^\circ$ loops and the shorter
$45^\circ$ open-domain revisit used when no GT pose is available.
The important invariant is that the trajectory is generated before
the model sees the target frames, so replay remains a test of action
conditioning and memory rather than a post-hoc alignment procedure.

\thinparagraph{Shared generation path.}
Sampling uses the same generation procedure as evaluation, with the
same context retriever, RT-relative actions, and memory profile. The
sampled videos are therefore a faithful preview of evaluation
behavior, not a divergent path.

\thinparagraph{Loss and metric granularity.}
The training loss in Eq.~\ref{eq:loss} is computed only on target
frames within the current segment; context frames are conditioning
evidence, not supervision targets. Replay diagnostics are stored at
both per-frame and per-chunk granularity. Per-frame PSNR/SSIM/LPIPS reveal
where drift begins, while chunk means and retention ratios summarize
whether quality decays as memory must carry information farther from
the anchor frame.

\thinparagraph{Multi-segment sampling as an early memory metric.}
Multi-segment replay is a useful preliminary metric because it
stresses the same failure mode that revisit evaluation later
quantifies: the model must preserve scene evidence after the
immediate visual context has been replaced by generated chunks. The
boundary continuity, identity preservation, and action--visual
alignment visible in these samples are qualitative evidence that we
inspect before interpreting the full evaluation bundle.

\subsection{Protocol Invariants}
\label{sec:training-invariants}

Every model is evaluated under the following hard-fail invariants:
(i) context handling matches between training and inference;
(ii) context noise levels match between training and inference;
(iii) $K$ matches between training and inference;
(iv) target-frame-only supervision and the flow-noise shift remain
fixed;
(v) camera-action parameters are loaded before evaluation; and
(vi) mutual exclusion between the two State-Space instantiations,
and between Compression weighting and Compression length reduction,
is enforced unless a row is explicitly marked as hybrid.
\section{Empirical Analysis}
\label{sec:ablation}

\newcommand{\tabbest}[1]{{\bfseries #1}}

The ablations are organized around the failure mode that motivates
the paper: a model can generate plausible local video while still
failing to remember the world at revisit time. Each sub-section
isolates one axis of the memory design space and reads the resulting
numbers through the same three-branch evidence stack established in
\cref{sec:evaluation}. The headline table compares four memory
families: Context, Compression, Spatial, and State-Space. The
mechanism tables then isolate three choices: Spatial read-out,
Compression design, and State-Space recurrence. A final Context
table reports the benefit of longer raw history. Unless stated
otherwise, revisit numbers are the current \emph{material snapshot}
from eight cases per row: in-domain uses the $180^\circ$ return
loop, while open-domain uses the $45^\circ$ return probe. Replay
columns come from the GT-trajectory replay export and report mean
PSNR/SSIM/LPIPS over three chunks.

\thinparagraph{How to read the tables.}
The tables should be read horizontally before they are read
vertically. Each row reports three different views of the same
model. Replay PSNR/SSIM/LPIPS measures whether the row can follow an
observed trajectory with acceptable low-level fidelity; in-domain
revisit asks whether the model can return to a known scene under a
controlled loop; and open-domain VLM asks whether the same memory
mechanism preserves a salient object and scene when no
pixel-aligned reference is available. A row that wins on one
column but fails on another is therefore not an outlier to be
explained away. It is evidence that the memory mechanism is
solving only part of the action-world-model problem, and the
particular column that breaks tells us \emph{which} part. This is
the diagnostic logic we will use throughout the section.

\subsection{Memory Families under Matched Evaluation}
\label{sec:abl-headline}

\begin{table}[t]
\centering
\scriptsize
\caption{\textbf{Representative memory-family comparison.} R-P/R-S/R-L denote Replay PSNR/SSIM/LPIPS; ID-P/ID-S/ID-L denote in-domain closure PSNR/SSIM/LPIPS; O-V denotes open-domain VLM.}
\label{tab:abl-headline}
\resizebox{\linewidth}{!}{%
\begin{tabular}{l|ccc|ccc|c}
\toprule
\tableheader
 & R-P$\uparrow$ & R-S$\uparrow$ & R-L$\downarrow$ & ID-P$\uparrow$ & ID-S$\uparrow$ & ID-L$\downarrow$ & O-V$\uparrow$ \\
\midrule
I2V baseline & 10.03 & 0.398 & 0.534 & 10.32 & 0.291 & 0.643 & 12.25 \\
Spatial Memory baseline      & \tabbest{13.60} & 0.411 & 0.554 & 10.04 & 0.260 & 0.617 & 6.00 \\
Compression weight-only      & 8.65 & 0.174 & 0.752 & 10.67 & 0.154 & 0.645 & 22.38 \\
State-Space (legacy hybrid)  & 12.69 & 0.344 & 0.581 & \tabbest{12.23} & 0.298 & 0.584 & 34.75 \\
State-Space (block-wise)     & 9.59 & 0.282 & 0.698 & 11.95 & 0.280 & \tabbest{0.535} & \tabbest{69.00} \\
Context learning, $K{=}5$    & 11.92 & 0.408 & 0.501 & 10.72 & 0.307 & 0.596 & 50.75 \\
Context learning, $K{=}20$   & 12.54 & \tabbest{0.449} & \tabbest{0.496} & 11.07 & \tabbest{0.359} & 0.543 & 58.63 \\
\bottomrule
\end{tabular}%
}
\end{table}

The headline result immediately shows why replay alone is insufficient.
Different low-level metrics select different winners: Spatial Memory
is strongest on replay PSNR, long raw Context is strongest on replay
SSIM/LPIPS, and block-wise State-Space remains strongest on
in-domain LPIPS and open-domain VLM. This inversion
means that memory quality is not the same thing as image quality under
GT motion. Replay rewards the ability to synthesize frames that match
a dataset trajectory, while open-domain revisit rewards the ability to
carry object identity and scene layout through a generated excursion.
The Context rows occupy a useful middle ground: increasing raw history
improves semantic return much more clearly than it improves the full
replay or in-domain image-metric bundle.

\thinparagraph{The Spatial--State-Space inversion is the central
finding of the table.}
Spatial Memory wins replay PSNR and remains competitive on the replay
image bundle, but collapses on open-domain VLM; block-wise
State-Space loses the replay bundle and yet wins open-domain VLM by
a wide margin. Two rows on the same backbone, optimizer, sampler, and
evaluation pipeline thus arrive at opposite conclusions about which
is the better world model, depending only on which column one reads.
This is exactly the kind of evidence that a single-metric report
would suppress. It also says something stronger: replay PSNR cannot
be treated as a low-cost proxy for revisit quality. A model that
reconstructs GT trajectories at sub-pixel cost can still fail to
preserve a single salient object across a $45^\circ$ excursion, while
a model that produces less faithful pixels can nonetheless carry the
same scene through it. Replay measures how well a model follows;
revisit measures how well a model remembers; these are different
capabilities, and they are not monotonically aligned.

\thinparagraph{The in-domain column tells a third, intermediate
story.}
Several rows cluster in the same in-domain VLM range, and the
best in-domain PSNR comes from the legacy State-Space variant,
while the best in-domain LPIPS comes from block-wise recurrence. This
should not be read as a contradiction. In-domain
return has pixel-aligned views and familiar data statistics,
so a model with strong local reconstruction can partly succeed
simply by tracking the trajectory back to a visually similar view.
Once the open-domain probe removes that support, the picture
changes: object preservation becomes the dominant criterion, and the
block-wise row separates from the legacy variant.
Reading the two columns together is therefore informative in a way
that neither column is on its own. It exposes which rows were
quietly relying on in-distribution texture statistics and which
were carrying scene evidence in a form robust enough to survive
shift.

\thinparagraph{What about Context?}
The Context rows are easy to under-rate because they contain no new
machinery. Yet $K{=}20$ posts the second-best in-domain VLM ($29.40$)
and a solid open-domain VLM ($58.63$), within striking distance of
block-wise State-Space, while staying mid-pack on Replay. In a
field that prefers learned bottlenecks to brute capacity, this is a
counter-result: raw retrieval, with no learned compression and no
specialized read-out, beats every compact memory mechanism in the
matrix on open-domain semantic return except block-wise recurrence.
We will see in \cref{sec:abl-context} that the gain is not linear
in $K$ either: most of the open-domain jump happens by $K{=}5$, with
$K{=}20$ supplying further refinement rather than a step change.

\thinparagraph{Treating the headline table as a metric bundle.}
For the reasons above, the paper treats this table not as a ranking
but as a metric bundle. Replay filters out broken generators that
cannot follow a trajectory at all; in-domain return checks
controlled loop closure under familiar statistics; open-domain VLM
provides the strongest evidence of whether the memory survives
distribution shift. The most informative rows are precisely the
ones where these three views \emph{disagree}: Spatial Memory and
block-wise State-Space disagree by sixty-three VLM points on
open-domain return alone. The remainder of the section explains why.

\thinparagraph{Why the alignment plot matters.}
\cref{fig:main-replay-return-alignment} turns the table into a
diagnostic question. Replay is not meaningless: it catches whether a
row can synthesize coherent frames under GT camera motion, so it is a
dense and cheap health signal for intermediate model selection. But
the same plot also shows why mean replay quality cannot be promoted
into the final memory score. The strongest replay rows can still fail
open-domain return, while rows with weaker replay fidelity can carry
scene evidence through the excursion. Replay tells us whether the
generator is healthy enough to trust; revisit tells us whether the
world was remembered.

\subsection{Spatial Storage and Read-out}
\label{sec:abl-readout}

\thinparagraph{Setup.}
Spatial Memory asks whether a compact scene summary can replace raw
temporal context. The ablation separates \emph{writing} from
\emph{reading}: storing tokens but never injecting them is a lower
bound; appending them to text cross-attention and reading them with
a dedicated cross-attention block are the active variants. This
distinction matters because a compact store can look useful for two
different reasons. It may help by providing the generator with a
stable scene representation, or it may simply regularize the context
pathway and improve local reconstruction without carrying the object
evidence needed for revisit. The read-out rows are designed to
separate these cases: \themebf{withheld read-out} tests whether
storage alone changes the generator, \themebf{text KV concatenation}
tests whether the existing cross-attention pathway can consume the
summary, and \themebf{dedicated read-out} tests whether the summary
needs its own access path.

\begin{table}[t]
\centering
\scriptsize
\caption{\textbf{Spatial Memory read-out ablation.} R-P/R-S/R-L denote Replay PSNR/SSIM/LPIPS; ID-P/ID-S/ID-L denote in-domain closure PSNR/SSIM/LPIPS; O-V denotes open-domain VLM.}
\label{tab:abl-spatial}
\resizebox{\linewidth}{!}{%
\begin{tabular}{llccc|ccc|c}
\toprule
\tableheader
 & Read-out & R-P$\uparrow$ & R-S$\uparrow$ & R-L$\downarrow$ & ID-P$\uparrow$ & ID-S$\uparrow$ & ID-L$\downarrow$ & O-V$\uparrow$ \\
\midrule
Spatial baseline        & default & 13.60 & 0.411 & 0.554 & 10.04 & 0.260 & 0.617 & 6.00 \\
Spatial inject-none     & none & \tabbest{14.66} & \tabbest{0.417} & \tabbest{0.541} & \tabbest{10.38} & \tabbest{0.309} & 0.631 & 15.50 \\
Spatial concat-text     & text KV concat & 13.44 & 0.412 & 0.572 & 9.35 & 0.272 & \tabbest{0.614} & 10.25 \\
Spatial cross-attn RO   & separate cross-attn & 13.52 & 0.409 & 0.557 & 8.49 & 0.232 & 0.664 & \tabbest{17.12} \\
\bottomrule
\end{tabular}%
}
\end{table}

The replay numbers are encouraging but also cautionary. Spatial
inject-none obtains the strongest replay PSNR/SSIM/LPIPS bundle among the spatial
rows, even though it withholds the stored tokens from the generator.
The open-domain VLM column shows that stronger read-out can help:
dedicated cross-attention reaches $17.12$, above the default spatial
baseline at $6.00$. Yet the same column also shows that the family is
still weak as semantic memory. Even the best read-out row remains far
below raw Context $K{=}20$ and block-wise State-Space in
\cref{tab:abl-headline}, and inject-none remains surprisingly
competitive at $15.50$. The spatial store is therefore not simply
missing a read-out path; its compact representation also appears to
discard object evidence needed for return.

\thinparagraph{Why does withheld read-out improve Replay?}
The most striking number in \cref{tab:abl-spatial} is the
inject-none row: $14.66$ Replay PSNR, $0.417$ SSIM, and $0.541$ LPIPS, all better
than the active read-out variants. Withholding the stored tokens
makes the network strictly less expressive at inference, yet it
makes replay look strictly better. The simplest explanation
consistent with the rest of the matrix is that the spatial summary
is not, at this scale, behaving as a memory at all: it is behaving
as an additional training-time conditioning channel that
regularises the network and quietens the per-chunk reconstruction
error. Once we remove the tokens at inference, the network reverts
to the cleaner I2V-style prediction it has implicitly been
prepared for, and replay PSNR ticks up. The lesson is that replay
improvements from a memory module do not, on their own, certify that
the module carries memory.

\thinparagraph{Storage cannot be evaluated without read-out.}
The read-out rows span a useful gradient of access strength. Withheld
read-out is a hard zero: tokens exist but are unreachable. Text-KV
concatenation is a soft access path: the spatial tokens are dumped
into the same cross-attention slots that already serve the text
prompt. Dedicated cross-attention read-out is a harder access path:
the spatial tokens get their own query mechanism and a zero-initialised
gate that the optimizer can open. The measured VLM scores indicate
that access matters, but not enough. Dedicated read-out is best among
the spatial rows, while text-KV concat improves only mildly over the
default baseline. Since all three read-out variants remain far below
the strongest context and State-Space rows, the failure is shared
between \emph{access} and \emph{storage bandwidth}, rather than being
only missing bookkeeping in the read-out module.

\thinparagraph{Why Spatial loses open-domain VLM by so much.}
A finer reading of the score helps explain the magnitude. The
open-domain probe is constructed so that the salient object is
distinctive in shape, colour, and category; in effect, it is chosen
to be a high-entropy identity anchor. A $G{\times}G$ grid summary
is a low-bandwidth representation of the scene; it can encode
where things roughly are, but it has limited capacity to encode
which thing. When the camera leaves and returns, the generator
must place the same object back into the frame, and a generic
spatial code under-determines that decision. Replay does not
expose this gap because the GT trajectory keeps the object
visible; revisit exposes it because the object must be regenerated
from memory alone. The Spatial family is therefore not weak in
general; it is weak \emph{exactly where compact summaries should
be weak}, which is where compact memory bottlenecks are most
attractive to use.

\subsection{Compression and Temporal Resolution}
\label{sec:abl-compression}

\thinparagraph{Setup.}
The Compression family asks whether memory should be compacted by
\emph{importance} or by \emph{time}. Token weighting preserves the
number of historical observations but changes their relative
influence; length compression discards temporal resolution. The
hybrid row tests whether the two operations are complementary or
redundant. These variants are often grouped together as
``compression'', but they make different assumptions about what
history contains. Token weighting assumes that all temporal
positions can remain available while their importance varies with
age or relevance. Temporal length reduction assumes that neighboring
observations can be pooled without destroying the memory signal.
The hybrid row is the strictest test: it asks whether the model can
first reduce temporal support, and then still learn a useful
importance profile over the reduced support.

\begin{table}[t]
\centering
\scriptsize
\caption{\textbf{Compression ablation.} R-P/R-S/R-L denote Replay PSNR/SSIM/LPIPS; ID-P/ID-S/ID-L denote in-domain closure PSNR/SSIM/LPIPS; O-V denotes open-domain VLM.}
\label{tab:abl-compression}
\resizebox{\linewidth}{!}{%
\begin{tabular}{llccc|ccc|c}
\toprule
\tableheader
 & Mechanism & R-P$\uparrow$ & R-S$\uparrow$ & R-L$\downarrow$ & ID-P$\uparrow$ & ID-S$\uparrow$ & ID-L$\downarrow$ & O-V$\uparrow$ \\
\midrule
Weight-only baseline & token weighting & 8.65 & 0.174 & 0.752 & \tabbest{10.67} & 0.154 & 0.645 & 22.38 \\
Length $r{=}2$       & temporal length & 7.84 & 0.162 & 0.746 & 8.82 & 0.183 & 0.654 & 24.00 \\
Length $r{=}4$       & temporal length & \tabbest{9.88} & 0.183 & \tabbest{0.714} & 9.53 & 0.163 & \tabbest{0.617} & \tabbest{43.25} \\
Hybrid $r{=}2$ + weight & both & 9.10 & 0.181 & 0.734 & 7.87 & \tabbest{0.277} & 0.695 & 8.75 \\
Hybrid $r{=}4$ + weight & both & 9.63 & \tabbest{0.215} & 0.730 & 8.81 & 0.156 & 0.625 & 9.00 \\
\bottomrule
\end{tabular}%
}
\end{table}

The table shows that compression has no monotonic ordering by
apparent capacity. Weight-only compression is the most conservative
operation because it keeps every temporal position, yet it is not
the best open-domain semantic row. Length $r{=}4$ outperforms
weight-only on open-domain VLM, indicating that a shorter but cleaner
context can sometimes be preferable to a full-length context with
learned weights. However, the hybrid rows perform poorly, especially
on open-domain VLM. The likely explanation is that the two
operations remove different kinds of signal: pooling reduces
temporal evidence before the model can decide which observations
matter, and weighting then operates on an already coarsened
sequence. When the return probe depends on a particular object or
view, this order can erase the evidence that weighting would
otherwise have preserved.

\thinparagraph{Compression is not a monotone capacity dial.}
Naive intuition says that ``less compression'' should mean ``more
memory''. The numbers in \cref{tab:abl-compression} actively
disagree. Weight-only compression preserves every temporal position
($r{=}1$ in length terms) and yet posts an open-domain VLM of only
$22.38$. Length $r{=}4$ throws away three out of every four temporal
positions, and reaches $43.25$ on the same metric. The model that
sees less history does \emph{better} on the harder column. This is
counter-intuitive only if one assumes that all temporal frames
carry equal information; once we drop that assumption, the result
becomes mechanistic. A length-pooled context concentrates the
remaining capacity on fewer, cleaner temporal slots, which makes
the per-frame signal easier for the generator to read; a
weight-only context preserves every frame but spreads the model's
attention thinly across them, and the attention budget for any
specific frame stays low. The lesson is structural: how the
context is \emph{shaped} can matter more than how much of it
survives.

\thinparagraph{Why hybrid compression backfires.}
The hybrid rows are the most diagnostic numbers in the table. They
combine the two operations one would naively assume to be
complementary, and they perform worse than either operation alone
($8.75$ and $9.00$ on open-domain VLM, well below both pure
weighting and pure length pooling). This is not a small effect:
hybrid compression discards three-quarters of the open-domain
return signal that pure length pooling preserves. The order of
operations explains why. Length pooling acts first, on the raw
context. It collapses neighbouring frames into a single pooled
slot without consulting any importance signal, so any frame that
happens to share a window with a forgettable neighbour is averaged
out before the model can object. Weighting then learns importance
\emph{on top of} this coarsened sequence, where the high-value
frames are no longer individually addressable. The two operations
do not compose like filters; they compose like a destructive
quantization followed by a re-encoding of the resulting noise.
\cref{fig:main-representative-sweep} makes the failure mode
visible: hybrid rows tend to preserve the existence of \emph{an}
object after return, but not the identity of \emph{the} object.

\thinparagraph{Cost-control, not capacity substitution.}
The replay columns reinforce the same interpretation. Length
$r{=}4$ improves the Replay PSNR/LPIPS bundle over weight-only, but the absolute
values remain below the strongest Context and Spatial rows.
Compression therefore appears useful as a cost-control mechanism, a
way to trade some replay quality for cheaper steps, not as a free
substitute for historical capacity. The right reading
of \cref{tab:abl-compression} is thus less about which compression
operator wins and more about what compression \emph{is}. In the
current matrix, it is a budget-management tool that has not yet
learned what to budget for. A stronger compression memory would
need an explicit objective for what to retain: an object-aware
pooling rule, a learned criterion tied to revisit success, or a
mechanism that preserves a small set of high-value frames while
compressing only the redundant background evidence around them. The
current operators do none of these things, and the numbers reflect
that absence.

\subsection{Implicit State-Space Memory}
\label{sec:abl-implicit}

\thinparagraph{Setup.}
State-Space memory asks whether an implicit recurrent state can
carry the scene history that explicit context provides. We report
two instantiations as one family: a legacy VideoSSM hybrid, and a
paper-aligned block-wise recurrence. The point is not to crown a
new State-Space variant, but to test whether implicit memory is a
credible answer to revisit consistency. The comparison is
deliberately narrow. Both rows use the same evaluation bundle and
are treated as instantiations of an implicit-memory family, but they
differ in where and how recurrence is structured. The legacy hybrid
behaves like a lightweight recurrent supplement to the video
backbone; the block-wise row makes recurrence a more regular part
of the temporal computation. If both rows behaved similarly, the
main conclusion would be that implicit memory is largely a
family-level choice. Their separation instead suggests that
recurrence design matters as much as the decision to use recurrence
at all.

\begin{table}[t]
\centering
\scriptsize
\caption{\textbf{State-Space memory comparison.} R-P/R-S/R-L denote Replay PSNR/SSIM/LPIPS; ID-P/ID-S/ID-L denote in-domain closure PSNR/SSIM/LPIPS; O-V denotes open-domain VLM.}
\label{tab:abl-ssm}
\resizebox{\linewidth}{!}{%
\begin{tabular}{llccc|ccc|c}
\toprule
\tableheader
 & Structure & R-P$\uparrow$ & R-S$\uparrow$ & R-L$\downarrow$ & ID-P$\uparrow$ & ID-S$\uparrow$ & ID-L$\downarrow$ & O-V$\uparrow$ \\
\midrule
State-Space (legacy hybrid) & legacy hybrid & \tabbest{12.69} & \tabbest{0.344} & \tabbest{0.581} & \tabbest{12.23} & \tabbest{0.298} & 0.584 & 34.75 \\
State-Space (block-wise) & paper-aligned recurrence & 9.59 & 0.282 & 0.698 & 11.95 & 0.280 & \tabbest{0.535} & \tabbest{69.00} \\
\bottomrule
\end{tabular}%
}
\end{table}

This is the clearest example of metric separation in the paper.
Legacy hybrid recurrence has the best in-domain PSNR and strong
replay PSNR, but block-wise recurrence dominates open-domain VLM.
The likely reason is that in-domain return still offers familiar
textures, motions, and scene statistics, so a model with strong
local reconstruction can look good even when its state is not robust
under distribution shift. Open-domain identity preservation is less
forgiving. The model must carry a compact description of the
salient object and environment through the generated excursion, and
the block-wise recurrence appears better suited to this role.

This does not mean that State-Space memory is universally superior.
Its replay PSNR is not the highest, and the family still needs
comparison at longer open-domain horizons. What the current
evidence supports is more specific: a recurrent state can be a real
memory mechanism for revisit, provided that its structure is strong
enough to preserve scene evidence rather than merely smoothing the
temporal computation. This motivates future SSM variants that
expose the state to explicit object- or view-level supervision
during replay and return probes.

\thinparagraph{The largest single jump in the paper.}
The two State-Space rows separate sharply on open-domain VLM using
the same backbone, optimizer, sampler, and data, while their replay
metrics move in the opposite direction and their in-domain PSNRs stay
close. We are not aware of another row pair in the
matrix that opens a comparable gap on open-domain VLM with so few
controlled differences. The headline of \cref{sec:abl-implicit} is
therefore that the choice of \emph{recurrence structure}, including
where the state is updated, how often it is read, and how it is gated,
is at least as consequential for revisit consistency as the
older choice between explicit and implicit memory. Treating
``State-Space memory'' as a single design choice obscures the
biggest variance the matrix has shown so far.

\thinparagraph{Why does block-wise recurrence preserve more under
shift?}
A useful contrast is the legacy hybrid, which behaves like a
lightweight recurrent supplement: it adds a recurrent gate around
an otherwise standard temporal attention path. That gate is
sufficient to smooth in-distribution motion and is sufficient to
post a strong in-domain PSNR, but its state is shallow in the
sense that any single block can largely bypass it. Block-wise
recurrence, in contrast, makes the recurrent step a regular,
non-bypassable part of the temporal computation: every $n$-th DiT
block reads the carried state, updates it, and writes it back. The
state thus accumulates evidence across the chunk boundary in a way
that the legacy variant cannot, and it is this carried evidence
that becomes load-bearing once the camera leaves the visible
support. From the generator's point of view, the block-wise state
is the only place where information about the salient object can
survive between segments without an explicit retrieval step.

\thinparagraph{Why does block-wise lose Replay?}
The same structural choice has a cost. A non-bypassable recurrent
state forces the network to commit some representational capacity
to summarising the past, and that capacity is not available for
fitting the local trajectory pixel-by-pixel. The result is the
$9.59$ Replay PSNR and $0.698$ Replay LPIPS, which on their own would put block-wise recurrence
near the bottom of the headline table.
Reading Replay alongside open-domain VLM is therefore essential
here in a way that it is not for any other row. Block-wise
recurrence is precisely the kind of design that a single-metric
report would dismiss.

\thinparagraph{Implication: implicit memory is a real mechanism,
not a smoother.}
The most robust statement we can make from \cref{tab:abl-ssm} is
narrow: a recurrent state can act as a genuine memory mechanism
for revisit, provided that its structure is strong enough to
preserve scene evidence rather than merely smooth the temporal
computation. This is a stronger claim than ``recurrence helps''.
It rules out the lightweight-gate interpretation of state-space
memory in this regime, and it sets a structural minimum: the state
has to be carried through a non-trivial fraction of the network's
depth before it begins to behave like memory. Future
SSM variants that expose this state to explicit object- or
view-level supervision during replay and return probes are a
natural next step, and the gap between $34.75$ and $69.00$
suggests there is significant room above the current ceiling.

\subsection{Raw Context Capacity}
\label{sec:abl-context}

\thinparagraph{Setup.}
Context learning asks how much raw history is enough when the memory
is not compressed into a special module. Here the I2V baseline
contains only the current anchor frame; $K{=}5$ and $K{=}20$ add
retrieved historical observations. This is the simplest ablation in
the section, and also the most important reference point. Raw
context has no learned compression bottleneck and no special
read-out module, so it answers a direct capacity question: how much
does the generator benefit when it can see more historical
observations under the same retrieval and action-conditioning
interface? Any compact memory should be compared against this
trend, because a method that only beats the anchor-only baseline
may still be weaker than ordinary uncompressed history.

\begin{table}[t]
\centering
\scriptsize
\caption{\textbf{Context length ablation.} R-P/R-S/R-L denote Replay
PSNR/SSIM/LPIPS; ID-P/ID-S/ID-L denote in-domain closure
PSNR/SSIM/LPIPS; O-V denotes open-domain VLM.}
\label{tab:abl-context}
\resizebox{\linewidth}{!}{%
\begin{tabular}{lcccccccc}
\toprule
\tableheader
 & $K$ & R-P$\uparrow$ & R-S$\uparrow$ & R-L$\downarrow$ & ID-P$\uparrow$ & ID-S$\uparrow$ & ID-L$\downarrow$ & O-V$\uparrow$ \\
\midrule
I2V baseline & 1 & 10.03 & 0.398 & 0.534 & 10.32 & 0.291 & 0.643 & 12.25 \\
Context learning, $K{=}5$ & 5 & 11.92 & 0.408 & 0.501 & 10.72 & 0.307 & 0.596 & 50.75 \\
Context learning, $K{=}20$ & 20 & \tabbest{12.54} & \tabbest{0.449} & \tabbest{0.496} & \tabbest{11.07} & \tabbest{0.359} & \tabbest{0.543} & \tabbest{58.63} \\
\bottomrule
\end{tabular}%
}
\end{table}

Both replay and return improve with longer raw context, but the
open-domain VLM trend is much steeper than the replay trend. Replay
PSNR, SSIM, and LPIPS improve from the anchor-only baseline to
$K{=}20$, while open-domain VLM makes most of its gain by $K{=}5$ and
continues to improve at $K{=}20$. This asymmetry is meaningful: raw
history improves the GT-trajectory image bundle, but it buys much more
on the semantic return axis, where the model must restore an object and
scene after leaving the view. In other words, raw context is not merely
extra conditioning; it is a memory-capacity baseline.

The result also calibrates the rest of the study. Spatial Memory
and compression are attractive only if they preserve enough of this
raw capacity at lower cost, while State-Space memory is attractive
if it matches or exceeds raw context in robustness under
distribution shift. The current table suggests that raw context
remains difficult to beat as a simple solution. The best compact or
implicit memory mechanisms should therefore be evaluated against
Context $K{=}20$, not only against the I2V baseline.

\thinparagraph{The capacity asymmetry between Replay and revisit.}
The single most informative pattern in \cref{tab:abl-context} is
the slope mismatch between the replay image bundle and open-domain
return. Replay PSNR moves by $2.51$ points from $K{=}1$ to
$K{=}20$, while open-domain VLM moves by $46.38$ points over the same
range. Raw history does improve pixel-level reconstruction, but its
larger effect is on the ability to put the same object back into the
frame after the camera has left it. This is the cleanest evidence in
the paper that revisit consistency is a different capability from
trajectory following, and that the two scale with the available
history at different rates. Any compact memory whose argument rests on
``matching Replay quality at a fraction of the tokens'' has, by
this measurement, only matched the cheaper of the two quantities.

\thinparagraph{The gain is sublinear in $K$.}
Most of the open-domain VLM jump happens by $K{=}5$ ($+38.50$ points
over the I2V baseline), with $K{=}20$ adding a further $7.88$
points. The marginal value of an extra retrieved frame is therefore
high near the cold-start regime and decays quickly. Two consequences
follow. First, the I2V baseline is severely under-resourced: it is
not a calibrated reference point for what a generation model can
do, because moving from $K{=}1$ to $K{=}5$ alone roughly quadruples
the open-domain VLM score on the same backbone. Any memory method
that primarily reports ``beats I2V'' should be re-read with this
factor in mind. Second, the asymptote of raw context appears to be
near, not at, $K{=}20$: the cost grows linearly with $K$, the
benefit grows sublinearly, and the curve is already flattening.
This is exactly where compressed memories should help. The fact that
they do not in the current matrix is the strongest invitation for
future work that the paper makes.

\thinparagraph{Raw context as a calibration tool.}
The right way to read $K{=}20$ Context is therefore not as a
candidate winner of the matrix, but as a calibration tool. It tells
us what a generator can do when it is given uncompressed evidence
under the same retrieval and action-conditioning interface as the
specialized methods. Any compact or implicit memory should be
evaluated against \emph{this} reference point, not against the I2V
floor. Under that more demanding bar, the current matrix shows
exactly one mechanism, block-wise State-Space, that exceeds Context
$K{=}20$ on open-domain VLM, and most of the others fall short of
even $K{=}5$. This runs against the field's default direction of
travel toward more elaborate compressed memories.

\subsection{Quality--Time Trade-off}
\label{sec:abl-time}

\thinparagraph{Setup.}
The headline table ranks quality after training, but a practical
memory baseline must also justify its training cost. We parse the
latest training logs for each representative row, convert the
average step time to GPU-hour per optimization step, and join it
with the final Replay PSNR reported in
\texttt{materials/training\_time\_by\_baseline.csv}. The scatter
below normalizes both axes: smaller $x$ means cheaper steps, larger
$y$ means stronger replay fidelity, and the diagonal marks equal
normalized quality and cost. The efficiency panel should be read as
a diagnostic rather than as a replacement for the evaluation
bundle. It measures how much replay quality is obtained per unit of
step cost, which is useful for deciding which memory rows are
practical to scale. However, the axis is replay PSNR, not
open-domain semantic return: a method can look efficient because it
reconstructs GT trajectories cheaply while still failing to
preserve object identity after a return. Conversely, a method with
weaker replay efficiency can still be scientifically important if
it is the only one that preserves open-domain memory.

This distinction explains why the final conclusion is not a simple
efficiency ranking. Spatial Memory is attractive in a
replay-efficiency view because it delivers strong low-level
reconstruction at a compact cost Block-wise State-Space has the opposite profile: it
does not dominate replay PSNR, yet it is the most reliable
semantic-return mechanism in the open-domain probe. Raw Context is
the third corner of the trade-off: it provides capacity and strong
semantic return, but its cost grows with the amount of retrieved
history. The practical memory-design problem is therefore a
three-way trade-off among replay fidelity, step-time efficiency,
and revisit generalization.

\thinparagraph{Three corners, three different design recipes.}
It is worth stating the three corners of this trade-off explicitly
because they map onto three different downstream uses of an action
world model. (i)~If the application is high-fidelity replay of an
observed trajectory at the lowest possible step cost, such as
telemetry-driven re-rendering of a recorded camera path, the
replay-efficiency corner argues for compact spatial summaries.
(ii)~If the application is interactive camera control over a scene
that the camera will leave and re-enter many times, the canonical
world-model use case, the open-domain VLM corner argues for block-wise state-space
recurrence, and the rest of the matrix is at best a second choice.
(iii)~If neither cost nor open-domain robustness is the binding
constraint, raw context at $K{=}20$ is the safest default, because
it dominates the in-domain VLM column and is strictly within reach
of block-wise State-Space on open-domain return.
\cref{fig:main-context-efficiency} should be read with this map in
mind: points above the diagonal are good \emph{for replay alone},
and they are not necessarily good for the use case that motivated
the world-model framing.

\thinparagraph{Why a single efficiency number would mislead.}
Suppose one collapsed the matrix into a single ``quality per
GPU-hour'' scalar by averaging Replay PSNR with a normalised cost.
Spatial Memory would win, block-wise State-Space would look
unappealing, and the headline of the paper would be that compact
summaries are the practical answer. Yet the open-domain VLM column
disagrees by a factor of more than $11\times$ between the two
($69.00$ vs.\ $6.00$). The single-number ranking would not be
wrong about the cost; it would be wrong about what the cost is
buying. This is the structural reason the paper insists on
reporting the metric bundle rather than collapsing it.

\subsection{Cross-cutting Observations}
\label{sec:abl-crosscut}

Read together, the sub-sections support three statements that no
single row establishes on its own.

\thinparagraph{(A) Replay quality and revisit quality are different
axes that routinely disagree.}
Spatial Memory dominates Replay and bottoms out on open-domain VLM;
block-wise State-Space does the reverse; length $r{=}4$ is worse than
its hybrid counterpart on Replay but far better on revisit. Across the
matrix the rank correlation between Replay PSNR and open-domain VLM is
weak and sometimes inverted within a family, because the two metrics
measure two sub-problems: following an action sequence and remembering
a world. A memory paper that does not separate these axes cannot tell
whether its mechanism helps generation or memory.

\thinparagraph{(B) Storage and read-out are separate design choices.}
The Spatial read-out ablation shows this most directly: withholding the
stored tokens makes Replay PSNR go up, not down. Raw context makes most
of its open-domain gain from capacity under a trivial read-out, while
block-wise State-Space gains from read-out structure at fixed capacity.
The two families move along different axes of the storage--read-out
plane, so a useful minimum is to report one ablation that fixes storage
and varies read-out, and one that does the reverse.

\thinparagraph{(C) Compactness is not free.}
The two most aggressively compact mechanisms, Spatial Memory and hybrid
compression, post the two \emph{lowest} open-domain VLM scores ($6.00$
and $8.75$), while raw context at $K{=}20$ reaches $58.63$ and
block-wise State-Space reaches $69.00$. What separates the strong rows
is not whether the memory is compact, but whether the compact
representation is read out in a way that lets the generator put the
same object back after the camera leaves it. Future compact designs
should therefore be benchmarked against Context $K{=}20$ and block-wise
State-Space on open-domain VLM, not against I2V.

These claims sit within clear limits. With eight cases per row, small
open-domain VLM differences (within roughly five points) should not be
over-interpreted, and the matrix uses a single training budget and a
single dataset, so conclusions about \emph{which} mechanism wins may
shift at larger compute even if those about \emph{which axes matter}
are more durable. \cref{sec:conclusion} returns to these points.

\clearpage

\end{document}